%% file: journal.tex
\documentclass[conference]{IEEEtran}
\pdfoutput=1
\IEEEoverridecommandlockouts
\usepackage{cite}
\usepackage{amsmath,amssymb,amsfonts}
\usepackage{algorithmic}
\usepackage{subfig}
\usepackage{textcomp}
\usepackage{xcolor}
\usepackage{nicefrac}       
\usepackage[pdftex]{graphicx}
\usepackage{multirow}
\usepackage{xspace}
\usepackage{enumitem}
\usepackage{booktabs}
\usepackage{tabu}
\usepackage{siunitx}

\newcolumntype{L}[1]{>{\raggedright\let\newline\\\arraybackslash\hspace{0pt}}m{#1}}
\newcolumntype{C}[1]{>{\centering\let\newline\\\arraybackslash\hspace{0pt}}m{#1}}
\newcolumntype{R}[1]{>{\raggedleft\let\newline\\\arraybackslash\hspace{0pt}}m{#1}}

\def\BibTeX{{\rm B\kern-.05em{\sc i\kern-.025em b}\kern-.08em
    T\kern-.1667em\lower.7ex\hbox{E}\kern-.125emX}}

\newcommand{\system}{RAD\xspace}
\newcommand{\systemVoting}{RAD Voting\xspace}
\newcommand{\systemActive}{RAD Active Learning\xspace}
\newcommand{\systemActiveLimit}{RAD Active Learning Limited\xspace}
\newcommand{\systemSlimmed}{RAD Slimmed\xspace}

\author{

   \IEEEauthorblockN{Zilong Zhao\IEEEauthorrefmark{1}\thanks{$^1$ This work has been partly supported by the IRS (Initialtive de Recherche Strat\'egique) program DATE.},  Robert Birke\IEEEauthorrefmark{2}, Rui Han\IEEEauthorrefmark{3}, Bogdan Robu\IEEEauthorrefmark{1}, Sara Bouchenak\IEEEauthorrefmark{4}, Sonia Ben Mokhtar\IEEEauthorrefmark{4}, Lydia Y. Chen\thanks{$^2$ This work has been partly funded by the Swiss National
Science Foundation NRP75 project $407540\_167266$ and TU Delft technology fellowship.}\IEEEauthorrefmark{5} }
    \IEEEauthorblockA{\IEEEauthorrefmark{1}Univ. Grenoble Alpes, France
    \{zilong.zhao, bogdan.robu\}@gipsa-lab.fr}    
    \IEEEauthorblockA{\IEEEauthorrefmark{2}ABB Corporate Research, Switzerland
    \{robert.birke\}@ch.abb.com}
     \IEEEauthorblockA{\IEEEauthorrefmark{3}Beijing Institute of Technology
    \{hanrui\}@bit.edu.cn}
    \IEEEauthorblockA{\IEEEauthorrefmark{4}INSA Lyon, France
    \{sara.bouchenak, sonia.benmokhtar\}@insa-lyon.fr}
    \IEEEauthorblockA{\IEEEauthorrefmark{5}TU Delft, Netherlands
    \{y.chen-10\}@tudelft.nl}

}

\title{RAD: On-line Anomaly Detection for Highly Unreliable Data}

\begin{document}

\maketitle

\input{Abstract}
\input{Introduction}
\input{Background}

\input{Framework}
\input{Evaluation}
\input{RelatedWork}
\input{Conclusion}

\bibliographystyle{abbrvname}
\bibliography{journal}

\end{document}

%% file: Abstract.tex
\begin{abstract}

Classification algorithms have been widely adopted to detect anomalies for various systems, e.g.,~IoT, cloud and face recognition, under the common assumption that the data source is clean, i.e.,~features and labels are correctly set.
However, data collected from the wild can be unreliable due to careless annotations or malicious data transformation for incorrect anomaly detection.
In this paper, we present a two-layer on-line learning framework for robust anomaly detection (\system) in the presence of unreliable anomaly labels, where the first layer is to filter out the suspicious data, and the second layer detects the anomaly patterns from the remaining data. To adapt to the on-line nature of anomaly detection, we extend \system with additional features of repetitively cleaning, conflicting opinions of classifiers, and oracle knowledge.  We on-line learn from the incoming data streams and continuously cleanse the data, so as to adapt to the increasing learning capacity from the larger accumulated data set. Moreover, we explore the concept of oracle learning that provides additional information of true labels for difficult data points. 
We specifically focus on three use cases, (i) detecting 10 classes of IoT attacks, (ii) predicting 4 classes of task failures of big data jobs, (iii) recognising 20 celebrities faces.
Our evaluation results show that \system can robustly improve the accuracy of anomaly detection, to reach up to 98\% for IoT device attacks (i.e.,~+11\%), up to 84\% for cloud task failures (i.e.,~+20\%) under 40\% noise, and up to 74\% for face recognition (i.e.,~+28\%) under 30\% noisy labels. The proposed \system is general and can be applied to different anomaly detection algorithms. 
 \end{abstract}



%% file: Introduction.tex
\section{Introduction}
\label{sec:Introduction}

Anomaly detection is one of the core operations for enforcing dependability and performance in modern distributed systems~\cite{Xue:TNSM18:Ticket,DBLP:journals/tpds/PhamWTBTKI17}. Anomalies can take various forms including erroneous data produced by a corrupted IoT device or the failure of a job executed in a datacenter~\cite{Birke:TNSM16:Cloud,BirkeDSN2014,Zhao:DSN19}.

Dealing with this issue has often been done in recent art by relying on machine learning-based classification algorithms over system logs~\cite{Fang:2010,Giantamidis:2016} or backend collected data~\cite{DBLP:conf/ijcai/ZhangLSLH19,Huang:2017}. These systems often rely on the assumption of clean datasets from which the classifier learns to distinguish between data corresponding to a correct execution of the system from data corresponding to an abnormal execution of the latter (i.e., anomaly detection). As workloads at real systems are highly dynamic over time, it is even more challenging to predict anomalies that can not be easily distinguished from the system dynamics, compared to the systems with static workloads.

In this context, a rising concern when applying classification algorithms is the accessibility to a reliable ground truth for anomalies~\cite{Cerf:NIPsWS18:Duao}. Typically, anomaly data is manually annotated by human experts and hence the generation of anomaly labels is subject to quality variation, so-called noisy labels. For instance, annotating service failure types for data centers is done by operators.

However, standard machine learning algorithms typically assume clean labels and overlook the risk of noisy labels.  Moreover, recent studies point out the increasing dirty data attacks that can maliciously alter the anomaly labels to mislead the machine learning models~\cite{Kang:2018,Fan:2018,He:2017}. As a result, anomaly detection algorithms need to capture not only anomalies that are entangled with system dynamics but also the unreliable nature of anomaly labels.

Indeed, a strong anomaly classification model can be learned by incorporating a larger amount of datasets; however learning from data with noisy labels can significantly degrade the classification accuracy, even for deep neural networks, at a non-negligible computation resource~\cite{Vagin:2011, DBLP:conf/iclr/ZhangBHRV17}. 
Such a concern leads us to ask the following question:
how to build an anomaly detection framework that can robustly differentiate between true and noisy anomalies and efficiently learn anomaly classification models from a succinct amount of clean data. The immediate challenge of capturing the dynamics of data quality lies at the fact that label qualities are not directly observable but only via anomaly classification outcomes that in turn are coupled with the noise level of data labels.

We extend Robust Anomaly Detector (\system)~\cite{Zhao:DSN19}, a generic framework that continuously learns an anomaly classification model from streams of event logs or images that are subject to label noise.  The original design of \system is composed of two layers of learning models, i.e., a data label model and an anomaly classifier. The label model aims at differentiating the label quality, i.e., noisy v.s. true labels, for each batch of new data and only "clean" data points are fed in the anomaly classifier. The anomaly classifier predicts the event outcomes that can be in multiple classes of (non)anomalies, depending on the specific anomaly use case. In this extension, we derive three alternatives of \system, namely, voting, active learning and slimmed, which use additional information, e.g., opinions of conflicting classifiers and queries of oracles. Moreover, we iteratively update the prediction of historical windows such that the weak prediction can be continuously improved the latest model.

To demonstrate the effectiveness of \system, we consider three use cases, i.e.,~detecting 10 classes of attacks on IoT devices~\cite{meidan2018n}, predicting four types of task failures for big data processing cluster~\cite{reiss2011google,Rosa-DSN15} and recognising the 20 most abundant celebrity faces~\cite{facescrub:2014} from open datasets.
Our results show that \system can effectively and continuously cleanse the data, i.e., selecting data streams with clean labels, and result in better anomaly detection accuracy per additional included data stream, compared to classifiers without continuous data cleansing. Specifically, under 30\% noise, \system achieves up to 98.35\%, 84.72\% (comparing to 96.1\% and 80.92\% of no selection on dataset) for detecting IoT device attacks and predicting cluster task failures, respectively. If we implement \systemActive on cluster dataset with the same noise level, the final accuracy could reach to 88.1\%. For face image dataset, final accuracy of \systemSlimmed under 30\% noise achieves to 74.14\% (comparing to 46.47\% of no selection on dataset). Furthermore, our study also shows that \system is stable even when the noise is very strong. And if we don't have many clean data at beginning to pre-train the model, \systemActive and \systemActiveLimit could still perform very well from a very bad starting model.

The remainder of the paper is organized as follows.
Section~\ref{sec:ProblemStatement} describes the motivating case studies.
Sections~\ref{sec:Framework} and \ref{sec:Evaluation} present the proposed \system framework and the results of its experimental evaluation, respectively.
Section~\ref{sec:RelatedWork} describes the related work, and finally, Section~\ref{sec:Conclusion} draws our conclusions and lessons learned.

%% file: Background.tex
\section{Motivating case studies}
\label{sec:ProblemStatement}
\begin{figure*}[htb]
	\begin{center}
		\subfloat[Use case of IoT thermostat device attacks]{
			\includegraphics[width=0.60\columnwidth]{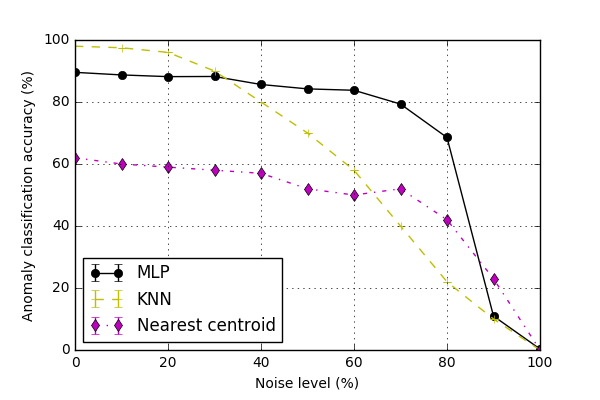}
			\label{fig:NoisyData-Classif-IoT}
		}		
		\hspace{\fill}
		\subfloat[Use case of Cluster task failures]{
			\includegraphics[width=0.60\columnwidth]{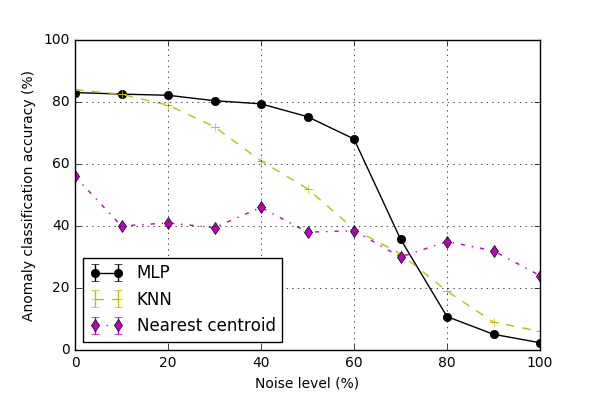}
			\label{fig:NoisyData-Classif-ClusterTasks}
		}
		\hspace{\fill}
		\subfloat[Use case of Face Recogniton]{
			\includegraphics[width=0.60\columnwidth]{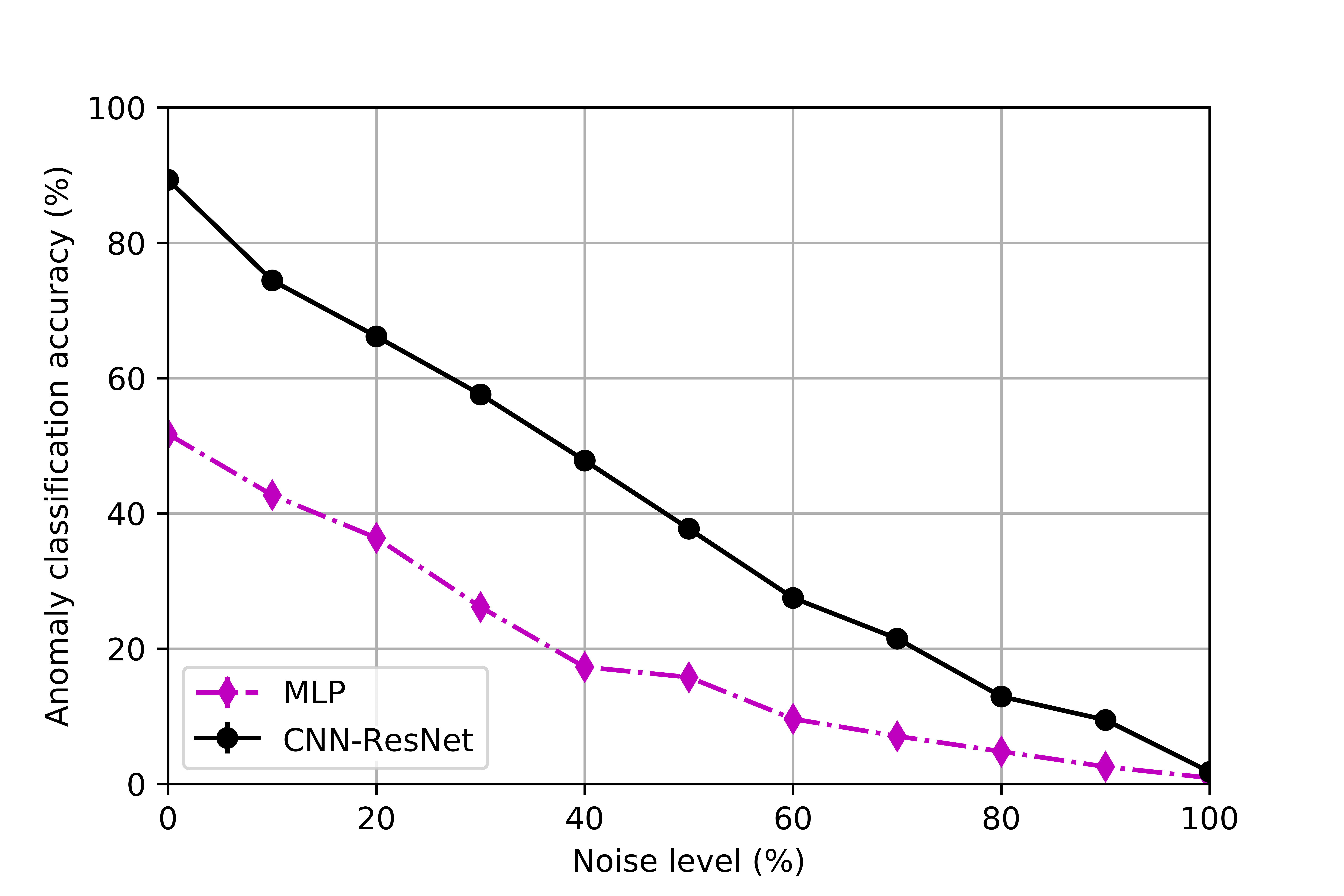}
			\label{fig:NoisyData-Classif-Facescrub}
		}
		\caption{Impact of noisy data on anomaly classification}
		\label{fig:NoisyData-Classif}
	\end{center}
\end{figure*}


To qualitatively demonstrate the impact of noisy data on anomaly detection, we use three case studies.

\begin{itemize}
\item Detecting \textbf{IoT device attacks} from inspecting network traffic data collected from commercial IoT devices~\cite{meidan2018n}. This dataset contains nine types of IoT devices which are subject to 10 types of attacks. Specifically, we focus on the Ecobee thermostat device that may be infected by Mirai malware and BASHLITE malware. 
Here we focus on the scenario of detecting and differentiating between 10 attacks. It is important to detect those attacks with high accuracies against all load conditions and data qualities.

\item Predicting \textbf{task execution failures} for big data jobs running at a Google cluster~\cite{reiss2011google,Rosa:TSC17:failurePrediction}. This trace contains a month-long jobs execution record from Google clusters.  Each job contains multiple tasks, which can be terminated into four different states: \textit{finish}, \textit{fail}, \textit{evict}, or \textit{kill}. The last three states are considered as anomaly states. To minimise the computational resource waste due to anomaly states, it is imperative to predict the final execution state of task upon their arrivals.

\item Recognizing \textbf{celebrity faces} from photos of the FaceScrub dataset~\cite{facescrub:2014}. The set is a collection of photos of celebrities roughly half female and half male. The task is to recognize faces by matching each photo to the identity of the celebrity shown on it. Here we focus on the face recognition of the 20 celebrities with the highest number of photos in the dataset totalling to 3.3K images. 
Faces are widely used in biometric identification systems in many security applications, e.g., access control. This makes the robustness of such systems critical. Furthermore, this image dataset is studied also because we want to show the broad applicability of our proposed framework.
\end{itemize}

The details about data definition, and statistics, e.g., number of feature and number of data points, can be found in Section~\ref{ssec:Datasets}. To recognize anomalies/faces in each use case, related studies have applied different machine learning classification algorithms, from simple ones, e.g., k-nearest neighbour (KNN), to complex ones, e.g., deep neural networks (DNN), under scenarios with different levels of label noise.
Here, we evaluate how the detection accuracy changes relative to different levels of noises. We focus on offline scenarios where we split the data in a training set affected by label noise and a clean evaluation set.

\subsection{Anomaly Detection}

Classification models are learned from 14,000 training records and evaluated on a clean testing dataset of 6,000 records. We specifically apply KNN, nearest centroid and multilayer perceptron (MLP) (a.k.a feed-forward deep neural networks) on both the IoT device attacks and the cluster task failures. Fig.~\ref{fig:NoisyData-Classif-IoT} and Fig.~\ref{fig:NoisyData-Classif-ClusterTasks} summarize the accuracy results.

One can see that noisy labels clearly deteriorate the detection results for both IoT attacks and task failures, across all three classification algorithms. For standard classifiers, like KNN and nearest centroid, the detection accuracy decays faster than MLP that is more robust to the noisy labels. Such an observation holds for both use cases. In IoT attacks, MLP can even achieve a similar accuracy as the case of no label noises, when 50\% of label classes are altered.

\subsection{Face Recognition}
For face recognition we use 2,639 images with varying degrees of label noise as the training set and 665 clean images as the testing set. Due to the particularity of image data, we use MLP and a specific CNN (Convolutional Neural Network) - ResNet (Residual neural Network) ~\cite{He2015DeepRL} as classification models. Fig.~\ref{fig:NoisyData-Classif-Facescrub} shows the accuracy results under the different label noise levels.

One can see, similar to the previous use cases, that label noise strongly affects the performance of both classifiers, although, the effect here is approximately linear. Moreover, ResNet performs better than MLP for this dataset under any noise level.

Above three experiments clearly show that under the presence of noisy label data, all the models are corrupted. The stronger the noise, the worse the model's accuracy. These cases motivate us to design the RAD framework and its extension. To resist the influence of noisy label data on learning process.



.
%

%

%
%

%% file: Framework.tex
\section{Design Principles of \system Framework}
\label{sec:Framework}



In this section, we will introduce the framework of \system along with its three extensions \systemVoting, \systemActive and \systemSlimmed. Different from other three, \systemSlimmed is specifically designed to deal with image dataset. All the symbols used to explain the designs are summarized in Table~\ref{tab:symboltable}.

\subsection{System Model}
\label{ssec:Definitions}



We consider a dataset that consists of several data instances.
Each data instance has $f$ features.
Each data instance belongs to a class $k$, where $k \in \mathcal{K}=\{1, \dots, \textit{K}\}$.
Data instances are either part of a pre-labeled dataset $\mathcal{D}$ with labels $Y$ used for training or non-labeled instances part of a dataset $\mathcal{P}$ used for inference.
Furthermore, a labeled data instance is either correctly labeled (i.e.,~clean data instance), or incorrectly labeled (i.e.,~noisy data instance). We use the indicator variable $\hat{q}$ to indicate clean $\hat{q}=1$ and dirty $\hat{q}=0$ labels. Wrong labels can stem from several reasons ranging from subjectivity, and data-entry errors, to malicious error injection.
The quality of a dataset $\mathcal{D}$ is measured as the percent of clean labeled data instances, denoted here as $\tilde{Q}$.

\begin{table}[h]
	\begin{center}
		\caption{Symbol description}
		\label{tab:symboltable}
		\begin{tabular}{L{1cm} C{6cm} }
			\toprule
			\textbf{Symbol}	& \textbf{Description}\\
			\midrule
			$\mathfrak{L}$ 	& label quality predictor					 \\
			$\mathfrak{C}$ & anomaly detection classifier					\\
			$\mathcal{D}_i$	&  $i$th training data batch 				\\
			$\mathcal{D}_i^*$	& $i$th  cleansed data batch from 	$\mathfrak{L}$			\\
			$\mathcal{P}_i$	& $i$th test data batch			\\
			$\hat{Y}_i$	& prediction of $i$th test data batch from 	$\mathfrak{C}$		\\
			$\tilde{Q}_i$ 	& percent of clean labeled data of $i$th batch				    \\
			
			$\mathcal{U}_i$	& "unclean" data of $i$th batch determined by 	$\mathfrak{L}$	 \\
			$\mathcal{U}_i^*$	& $i$th cleansed data batch from $\mathfrak{C}$			\\
			$\mathcal{S}_i$	& "unclean" data  of $i$th batch determined by $\mathfrak{C}$		\\
			$\mathcal{S}_i^*$	&   data with true label from Expert of $i$th batch	\\
			$\hat{p}$ & indicator of prediction, 1 for clean, 0 for dirty \\
			$\hat{q}$ & indicator of prediction, 1 for clean, 0 for dirty \\
			\bottomrule
		\end{tabular}
	\end{center}
\end{table}

Data instances arrive at the learning system continuously over time in batches.
$\mathcal{D}_i$ denotes the batch of labeled data arriving at time $t_i$ and having labels $Y_i$. In general we denote the time window with the subscript $i$.
We assume that a small initial batch of data instances $\mathcal{D}_0$ has only clean labels, that is $\tilde{Q}_0 = 100\%$.
Subsequent batches, include varying proportions of noisy labels, i.e~$ 0 < \tilde{Q}_i < 100\% , i > 0$.
For simplicity we consider arriving batches of equal size, $\forall \mathcal{D}_i, |\mathcal{D}_i| = N $, but not necessarily at regular times.

A classification request consists of a batch of non-labeled data instances $\mathcal{P}_i$ for which the classifier predicts the class $k$ of each data instance. At each batch arrival, the classification output $\hat{Y}_i$ is thus an array of the predicted classes for each non-labeled data instance. 

\begin{figure}[htb]
    \centering
    \includegraphics[width=\linewidth]{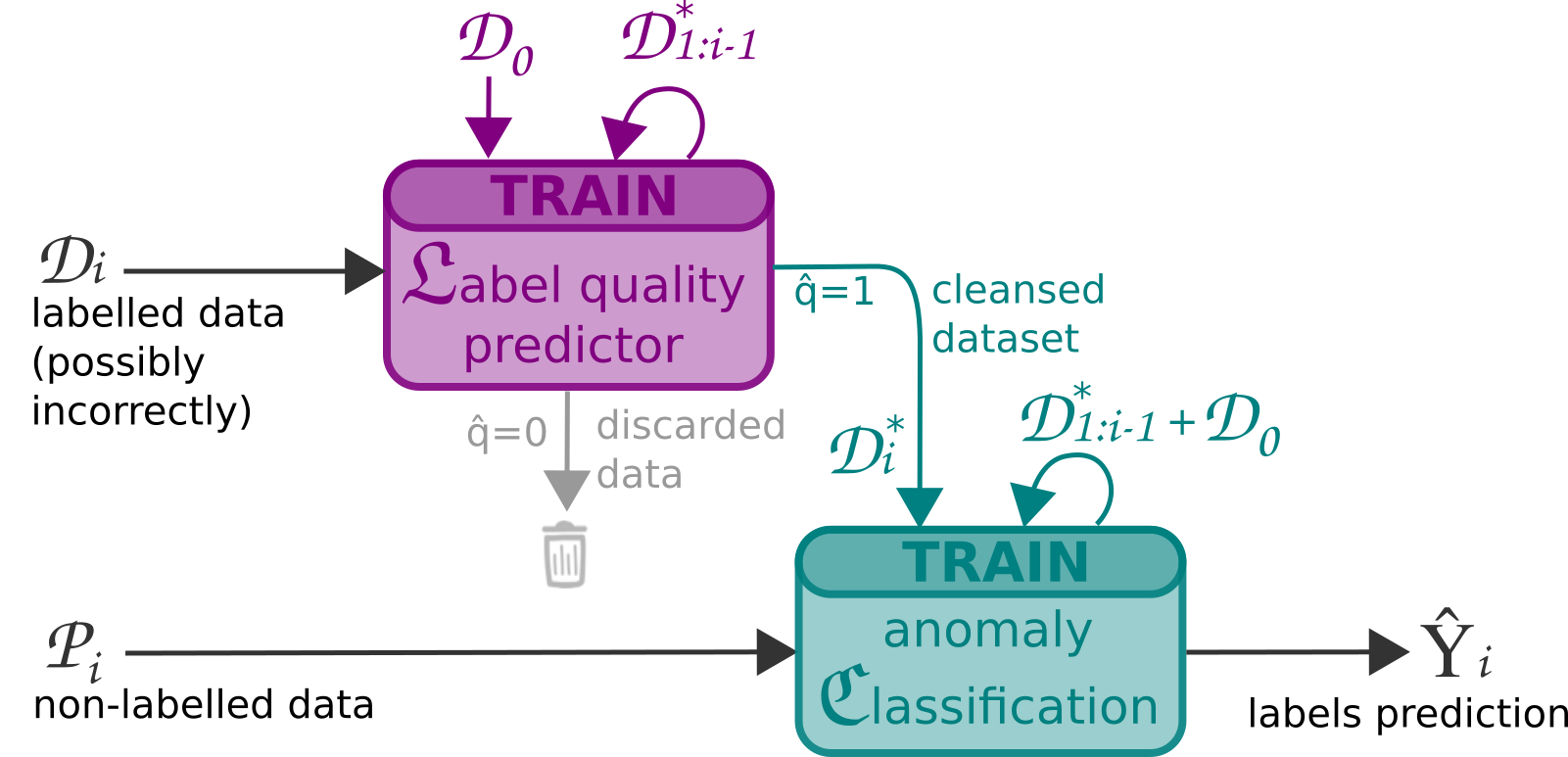}
    \vspace{0.5em}
    \caption{\system learning framework. Each block is a machine learning algorithm $\mathfrak{L}$ and $\mathfrak{C}$. Data used to train is represented by colored arrows from the top. The flowchart is iterated at every batch arrival with new labelled and unlabelled data coming in (black arrows on the left).  The labelled training data for $\mathfrak{C}$ is cleansed based on the label quality predicted by $\mathfrak{L}$. The incoming unlabelled data is classified by $\mathfrak{C}$ to generate the main output (black arrow on the right).
    }
    \label{fig:architecture}
\end{figure}

\subsection{Objectives and Overview of \system Framework}
\label{ssec:Overview}

We propose the \system learning framework. Its objective is threefold:

\begin{enumerate}

   	\item[(i)] Accurately learn a data model from noisy data.

	\item[(ii)] Continuously update the learned model based on new incoming data.

	\item[(iii)] Propose a general approach that caters to different machine learning algorithms, and different application use cases.

\end{enumerate}

Fig.~\ref{fig:architecture} describes the overall architecture of \system. \system comprises two main components. A label quality model $\mathfrak{L}$ aims at discerning clean labels from dirty labels and a classifier model $\mathfrak{C}$ targets the specific classification task at hand.

\subsubsection{Determining Data Noise}
\label{ssec:DataNoise}

The first component of \system aims to determining for each data instance in $\mathcal{D}$ if it is correctly or incorrectly labeled.
The objective of the label quality model is to select the most representative data instances to train a strong classifier model.
It solicits data instances with clean labels, avoiding the pitfall that the classifier overfits the noise.
\system uses supervised-learning algorithms to continuously train the label quality model from accumulated predicted clean data instances, which are highly correlated to a stronger classifier.

$\mathfrak{L}_i$ is the label quality model that is trained with data instances received up to time $t_i-1$, that is $\mathcal{D}_0 \dots \mathcal{D}_{i-1}$.
Upon the arrival of a new batch of data instances $\mathcal{D}_{i}$ at time $t_{i}$, we use the currently learned label quality model $\mathfrak{L}_{i}$ to predict the label quality $\hat{q}$ for each data instance in $\mathcal{D}_{i}$ by comparing the given $k$ and predicted class $\hat{k}^{\mathfrak{L}_{i}}$. If they coincide,  we consider the label as clean $q = 1$, otherwise as dirty $q = 0$.
Then we build $\mathcal{D^{*}}_{i}$ as the subset of data instances from $\mathcal{D}_{i}$ with $q = 1$ and discard the instances with $q = 0$.
We incorporate $\mathcal{D^{*}}_{i}$ into the existing training set for both the future label quality model $\mathfrak{L}_{i+1}$ and current classifier model $\mathfrak{C}_{i}$.


\subsubsection{Generic Approach to Handle Dynamic Data}
\label{ssec:GenericApproach}

The second component of \system is the dynamic data classifier $\mathfrak{C}$.
$\mathfrak{C}_{i}$ is trained on all the predicted clean data instances $\mathcal{D^{*}}$ received until time $t_i$, that is $\mathcal{D^{*}}_0 \dots \mathcal{D^{*}}_{i}$.
We assume that $\mathcal{D}_0$ contains only clean data instances to kickstart the framework and use the label quality model $\mathfrak{L}_{0} \dots \mathfrak{L}_{i-1}$ to cleanse $\mathcal{D}_1 \dots \mathcal{D}_{i}$ and produce $\mathcal{D^{*}}_1 \dots \mathcal{D^{*}}_i$.
Thus, the \system learning framework uses the batch-by-batch updated data label quality model to enrich the training data of the classification model.

\system follows a generic approach since the proposed classification framework can be used with any supervised machine learning algorithm, such as SVM, KNN, random forest, nearest centroid, DNN, etc.
Moreover, \system can be applied to a large spectrum of different applications where noisy data are collected and must be cleansed before used to train the classification model. Examples are the failure detection, attack diagnosis and face recognition illustrated in Section~\ref{sec:Evaluation}.

\subsection{Extensions to the Base \system}
\label{ssec:extensions}

The base framework can be extended in many different ways. Here we present three extensions addressing specific pitfalls and opportunities.

\subsubsection{Voting and History}
\label{ssec:GenericApproach}

\begin{figure}[htb]
	\centering
	\includegraphics[width=\linewidth]{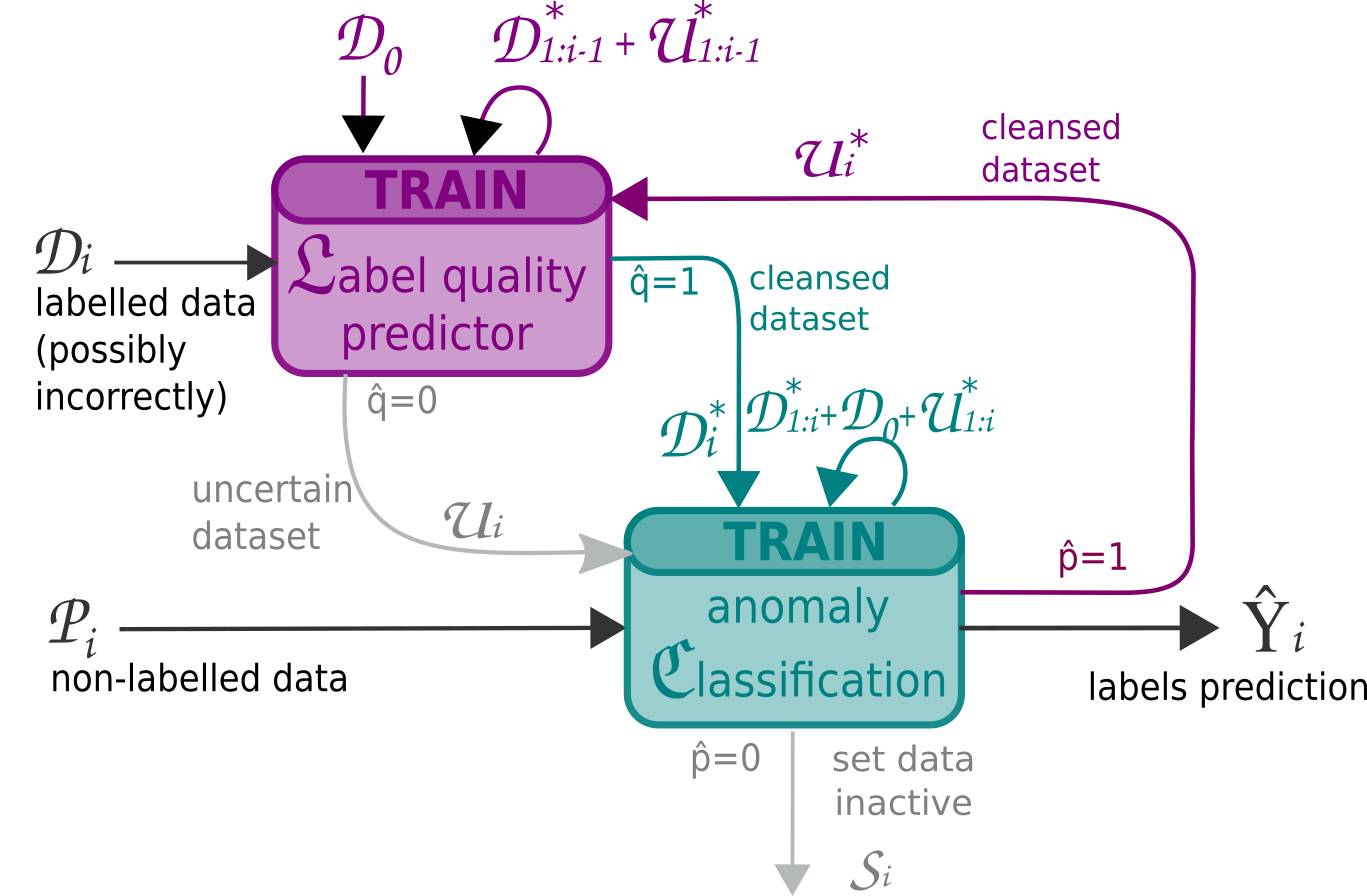}
	\vspace{0.5em}
	\caption{RAD voting correction.}
	\label{fig:architecture_voting}
\end{figure}

The base \system aims for separation of concerns with distinctive goals for the two models. However this approach biases the results towards the label quality model $\mathfrak{L}$. Hence we want the classifier model $\mathfrak{C}$ to also play a role in selecting clean data instances. We do this via the voting extension shown in
Fig.~\ref{fig:architecture_voting}.

Comparing to the base \system, predicted dirty labels having $\hat{q} = 0$ are not discarded by $\mathfrak{L}$ but passed to $\mathfrak{C}$ as uncertain data $\mathcal{U}$. Then the classifier $\mathfrak{C}$ is used to further cleanse the uncertain data to produce $\mathcal{U^*}$.
For each data instance in $\mathcal{U}$ we predict its class $\hat{k^{\mathfrak{C}}}$ using $\mathfrak{C}$ and looking for agreement with the given class $k$ and the class $\hat{k^{\mathfrak{L}}}$ predicted by $\mathfrak{L}$.
We add data instances to $\mathcal{U^{*}}$ if either $\hat{k^{\mathfrak{C}}}$ equals $k$, or if $\hat{k^{\mathfrak{C}}}$ equals $\hat{k^{\mathfrak{L}}}$. In the latter we replace the given class by the predicted class.
Then we incorporate both $\mathcal{D^{*}}$ and $\mathcal{U^{*}}$ into the training set for the classifier model $\mathfrak{C}$. We also send $\mathcal{U^{*}}$ to the label model $\mathfrak{L}$ to retrain for the next batch.

Batches of data instances not added to $\mathcal{U}_i^*$ at time $t_i$ are not immediately discarded but kept in a batch $\mathcal{S}_i$ of inactive data. The idea is that since the accuracy of the classifier improves over time (see Section~\ref{ssec:EvalDynamicData}), we can use the new classifier to re-evaluate old batches of inactive data and further increase the training data.
More in detail we maintain a list of the batches of inactive data $\mathcal{S}_i$ ordered by their size. After we finished training a new classifier we select the two biggest batches and re-process them via the voting system as before.

\subsubsection{Active Learning}

In \systemVoting we use the two models $\mathfrak{C}$ and $\mathfrak{L}$ to correct labels and increase the overall amount of data used for training aiming for improved  framework accuracy. However still not all data is used. To increase further the amount of training data we resort to active learning, i.e., we ask an expert for the true class of the data instances we are least certain about.

\begin{figure}[htb]
	\centering
	\includegraphics[width=\linewidth]{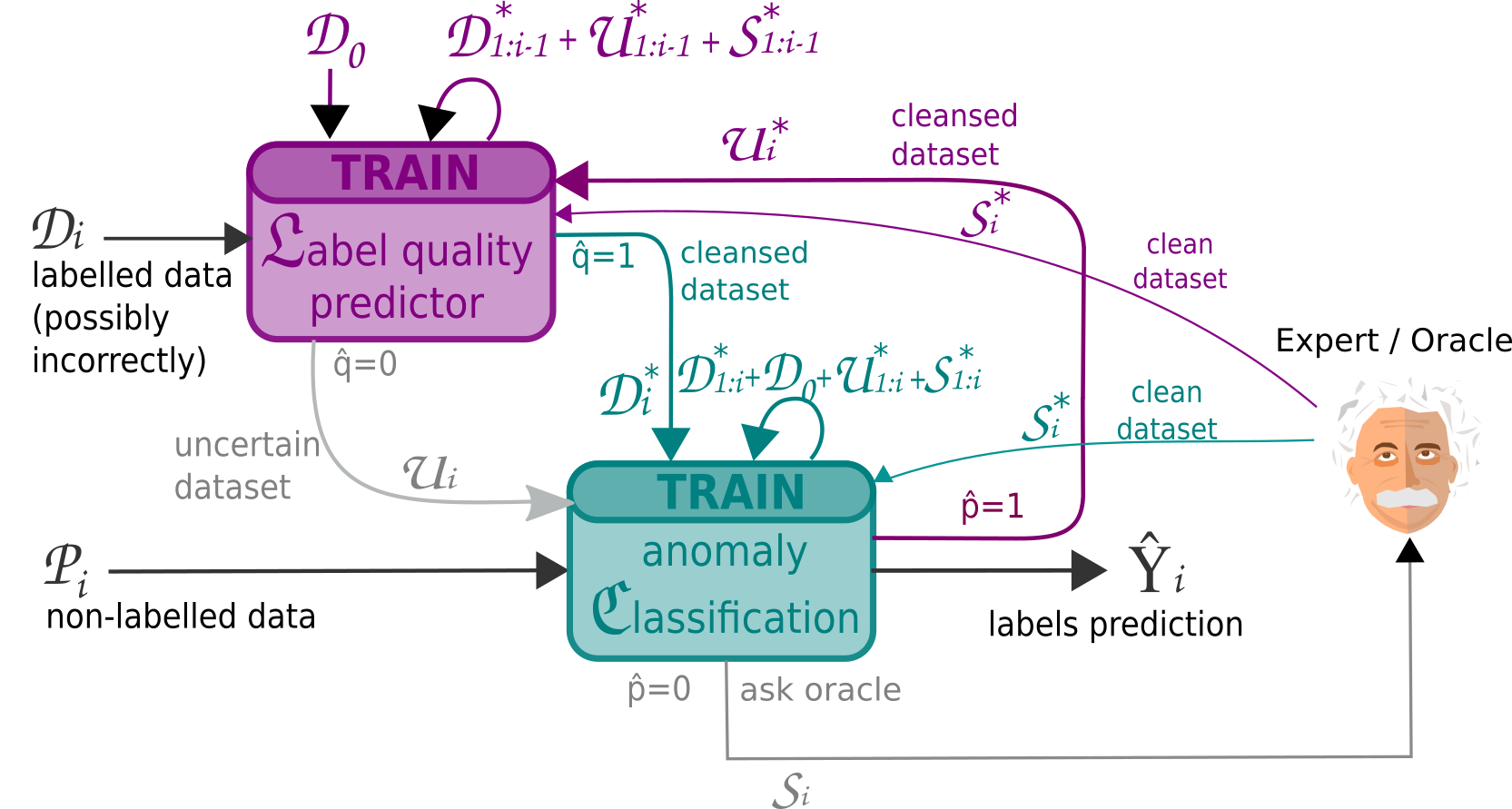}
	\vspace{0.5em}
	\caption{RAD - Active Learning.}
	\label{fig:architecture_oracle}
\end{figure}

Fig.~\ref{fig:architecture_oracle} shows the structure of \systemActive. The difference with the structure of \systemVoting is that in \systemActive we send the most uncertain data instances not to the inactive set $\mathcal{S}$ but to an oracle to ask for the true label. In \systemActive, potentially every data instance will be used to train $\mathcal{L}$ and $\mathfrak{C}$ and there is no inactive data anymore.
In reality, consulting an oracle for every single uncertain data instance might be too expensive. Hence we also consider \systemActiveLimit which additionally imposes a configurable limit on the number of queries to be asked to the oracle at each batch arrival. If the number of uncertain data instances exceeds this limit, we use random sampling. 

\subsubsection{\systemSlimmed}

The \system framework requires two models. Depending on the complexity of the models used the cost of training might be excessive. Especially in scenarios relying on complex deep neural networks, such as Convolutional Neural Networks (CNNs) for image classification, it might be to expensive and time consuming to train two models. To reduce the computational costs we propose a slimmed version of \systemActive named \systemSlimmed. The idea is to partially delegate the role of the label quality $\mathfrak{L}$ model to the oracle.

\begin{figure}[htb]
	\centering
	\includegraphics[width=\linewidth]{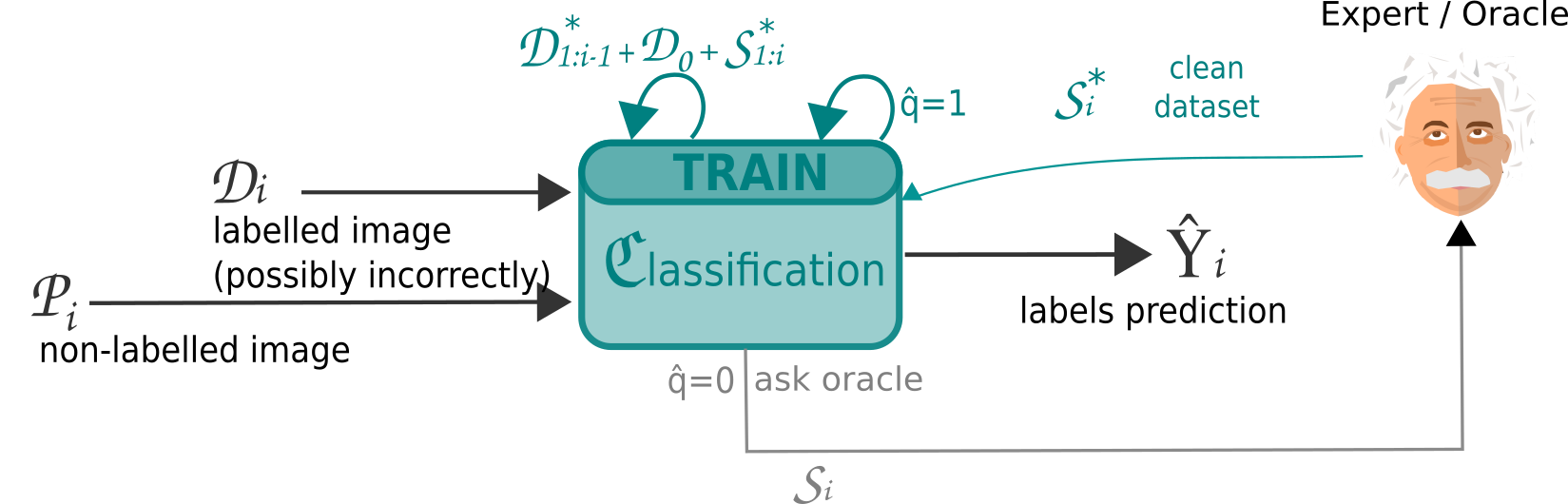}
	\vspace{0.5em}
	\caption{\systemSlimmed.}
	\label{fig:architecture-oracle-image}
\end{figure}

In \systemSlimmed new data batches arrive directly at the  $\mathfrak{C}$ model, see Fig.~\ref{fig:architecture-oracle-image}. For each data instance we compare the given label $k$ and to the predicted label $\hat{k^{\mathfrak{C}}}$. If they are the same we add to $\mathcal{D}^*$. If they differ we ask the oracle for the true label (possibly within a given query budget) and add it to $\mathcal{S}^*$. To train the model for the next data batch arrival, we use $\mathcal{D}^*$ as before plus $\mathcal{S}^*$ from the past two arrivals. The reasoning is that $\mathcal{S}^*$ stemming from an oracle has all correct labels and hence we want to double the learning from these data.

%% file: Evaluation.tex
\section{Experimental Evaluation}
\label{sec:Evaluation}


In this section, we implement \system , \systemVoting and \systemActive on IoT and Cluster datasets. Evolution of learning accuracy under 30\% and 40\% noise level are reported for all three frameworks. For \system , impact of noise level on final accuracy is discussed in section.~\ref{ssec:EvalNoiseImpact}. For \systemVoting , analysis on percentage of active and active-truth data changing over time is carried out in section.~\ref{ssec:SystemVoting}. \systemActive and its small update \systemActiveLimit are explained in section.  The Impact of size of initial data batch $\mathcal{D}_0$ on above frameworks are studied in section.~\ref{ssec:ImpactInitialization}.  To demonstrate the applicability of the framework to image dataset, \systemSlimmed is proposed, the results of \systemSlimmed with different limitations are reported in section.\ref{ssec:systemSlimmed}.

\subsection{Use Cases and Datasets}
\label{ssec:Datasets}

In order to demonstrate the general applicability of the proposed \system framework for anomaly detection, we consider the following three use cases:
(i)~Cluster task failures , (ii)~IoT botnet attacks and (iii)~Face recognition.
In our experiments, we use real data collected in cluster and IoT platforms and real celebrity face images.

The cluster task traces comprise data instances each corresponding to a task with 27 features capturing information related to static and dynamic system states, e.g. the task start/end times, the task resource utilisations, the hosting machine, etc. Each class is labeled based on its scheduling state. A detailed description of the features and labels can be found in~\cite{reiss2011google}. In particular, we are interested in the four possible termination classes: \textit{finish}, \textit{fail}, \textit{evict}, or \textit{kill}.
We filter out other classes. The resulting class distribution is dominated by successful tasks (\textit{finish}) 77.8\%, followed by \textit{kill} 22.0\%, \textit{fail} 0.2\%, and \textit{evict} $<$0.1\%.
Similar to~\cite{Rosa:TSC17:failurePrediction}, we aim to predict the task outcome to reduce the resource waste and improve the overall scheduling and system performance, e.g., in case of lack of resources and need to kill a task, help choosing the task with the least probability to succeed. We apply \system to continually train a noise-resistant model for better accuracy.


The IoT dataset comprises data instances describing 23 network packet-level statistics recursively computed over five different time scales totalling to 115 features. This traffic statistics are collected during normal operation, labeled as benign, or under one of ten different malicious attacks stemming from devices infected by either the {\em BASHLITE} or {\em Mirai} malware. Malicious traffic covers mainly scanning for vulnerable devices and various flooding attacks. The dataset provides traces collected at different IoT devices. More details are provided in~\cite{meidan2018n}. We aim to apply \system to build a noise-resistant model to categorize the attacks for post fact analysis, e.g., for threat assessment.

The FaceScrub~\cite{facescrub:2014} dataset is used for face recognition. Original FaceScrub contains more than 100,000 face images of 530 people, with about 200 images per person. Male and Female images are almost equal. We use a subset of 3.3K FaceScrub images to fit the limits of our compute resources. The 3.3K images cover the 20 people which have the highest number of images, 12 males and 8 females. FaceScrub images were retrieved from the Internet and are taken under real-world situations (uncontrolled conditions). We resize all images to 128*128 pixels. Name is the only annotation we use. The face recognition system has been widely used in security equipment.
We apply RAD Slimmed to FaceScrub dataset to show that our framework can help to build also robust face recognition models.



The main dataset characteristics are summarized in Table~\ref{tab:datasets}.

\begin{table}[htb!]
	\begin{center}
		\caption{Dataset description}
		\label{tab:datasets}
		\begin{tabular}{L{1.8cm} C{1.8cm} C{1.8cm} C{1.4cm}}
			\toprule
			\textbf{Use case}	& \textbf{Cluster task failures}	& \textbf{IoT device attacks}  & \textbf{FaceScrub}\\
			\midrule
			\#trainig data instances	& 60,000					& 33,000 & 2,639 \\
			\#test data instances	& 6,000					& 6,000 & 665\\
			\#classes $\textit{K}$ 		& 4 						& 11  & 20\\
			\#features $f$ 		& 27 						& 115 & 128*128\\
			data batch size 	& 600					    & 300   & 400\\
			initial batch $\mathcal{D}_{0}$ size & 6,000 &6,000&639\\
			\bottomrule
		\end{tabular}
	\end{center}
\end{table}


\subsection{Experimental Setup}
\label{ssec:ExperimentalSetup}

\begin{figure*}[ht]
	\centering
	\subfloat[With data noise level of 30\%]{
		\includegraphics[width=0.8\columnwidth]{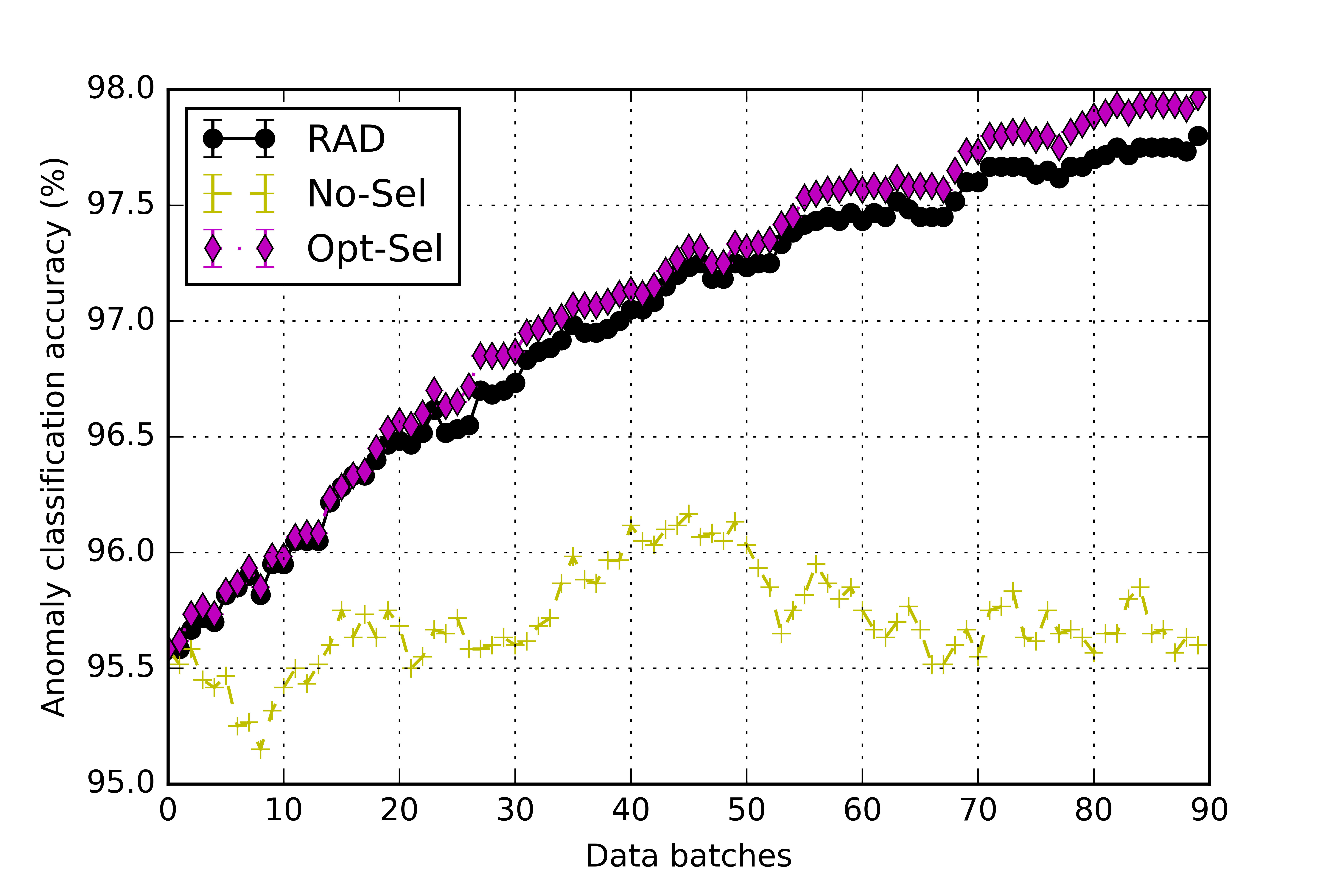}
		\label{fig:EvolOverTime-IoT-30}
	}
	\hfil
	\subfloat[With data noise level of 40\%]{
		\includegraphics[width=0.8\columnwidth]{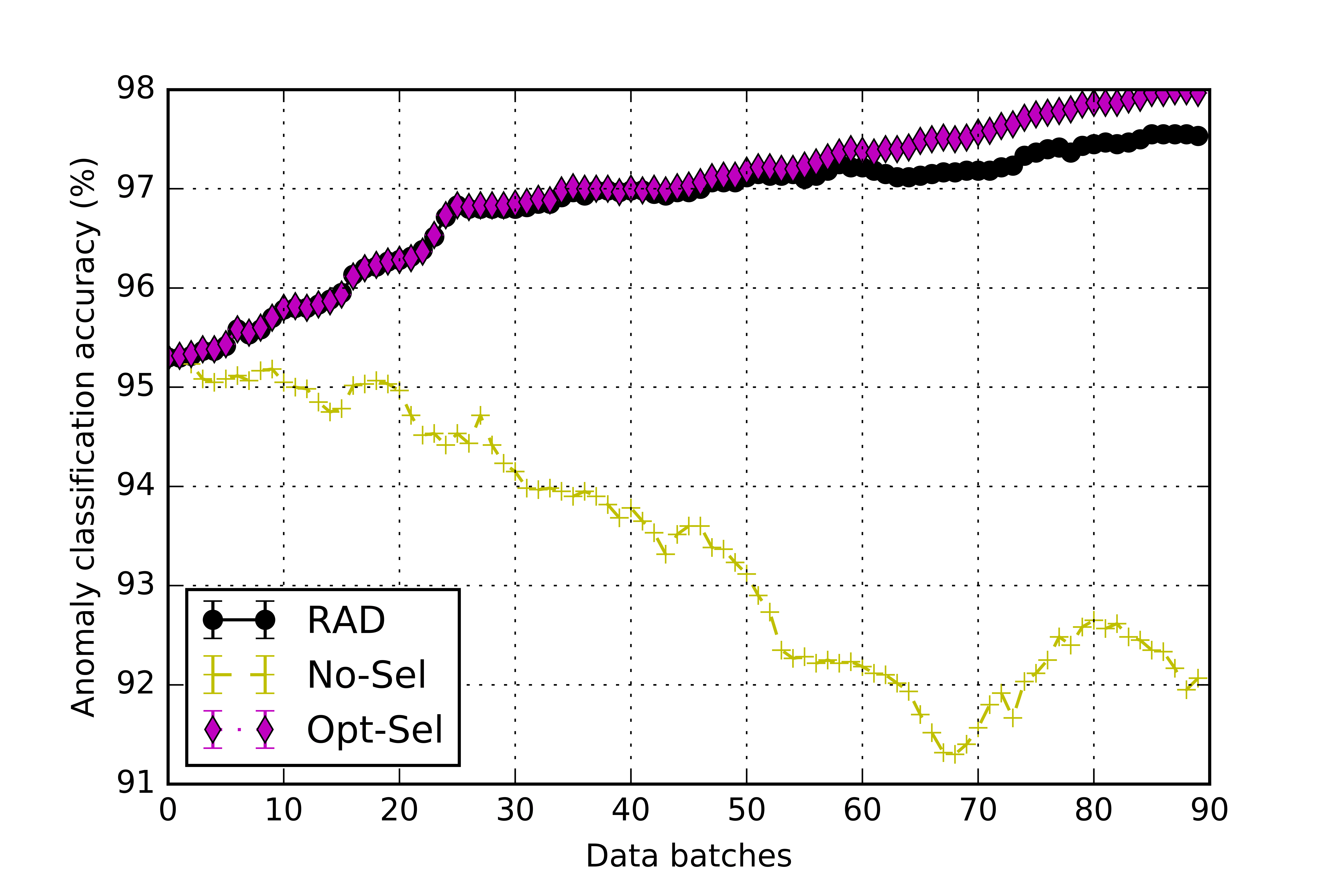}
		\label{fig:EvolOverTime-IoT-40}
	}
	\caption{Evolution of learning over time -- Use case of IoT thermostat device attacks}
	\label{fig:EvolOverTime-IoT}
\end{figure*}

\begin{figure*}[ht]
	\begin{center}
		\subfloat[With data noise level of 30\%]{
			\includegraphics[width=0.83\columnwidth]{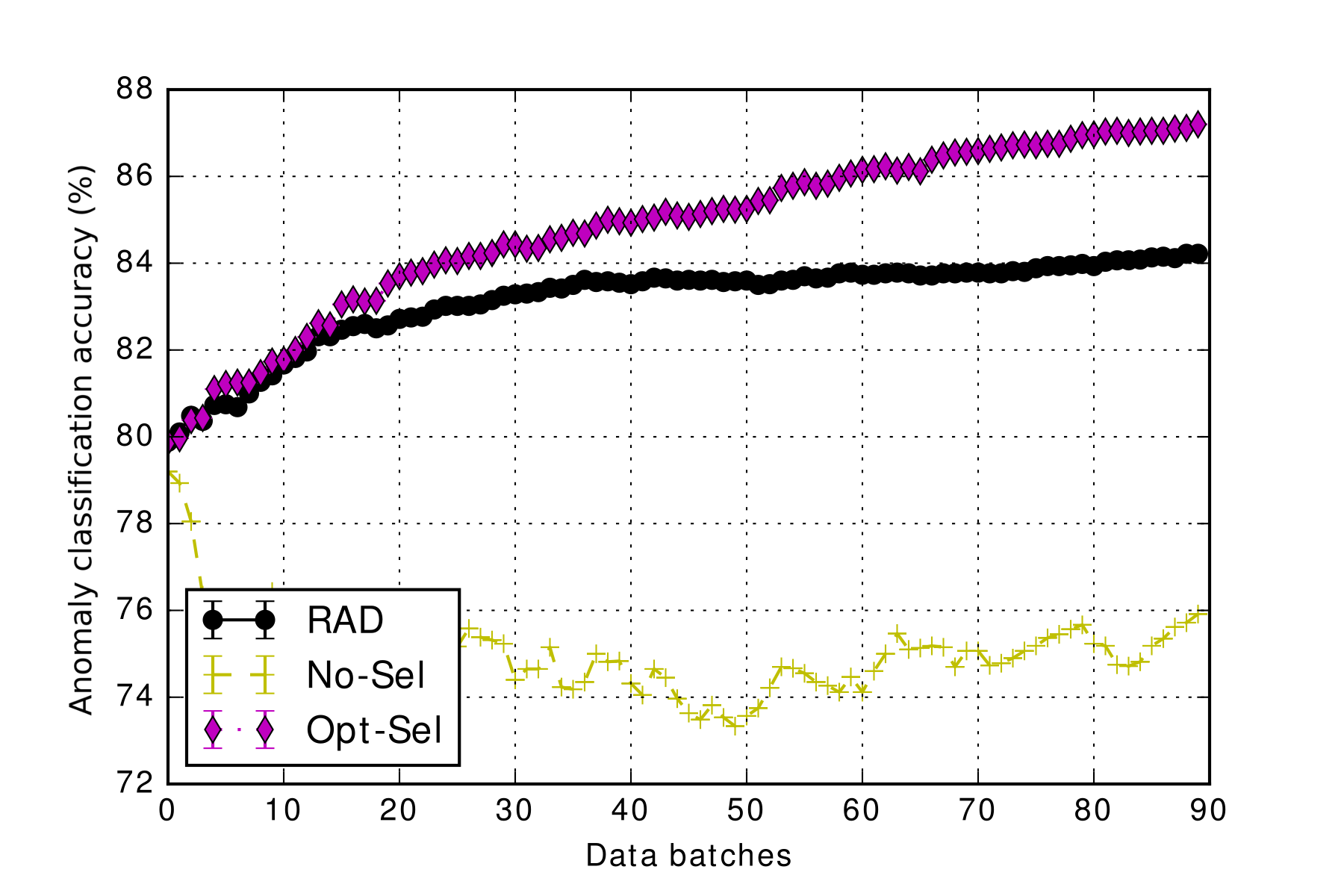}
			\label{fig:EvolOverTime-ClusterTasks-30}
		}
		\hfil
		\subfloat[With data noise level of 40\%]{
			\includegraphics[width=0.83\columnwidth]{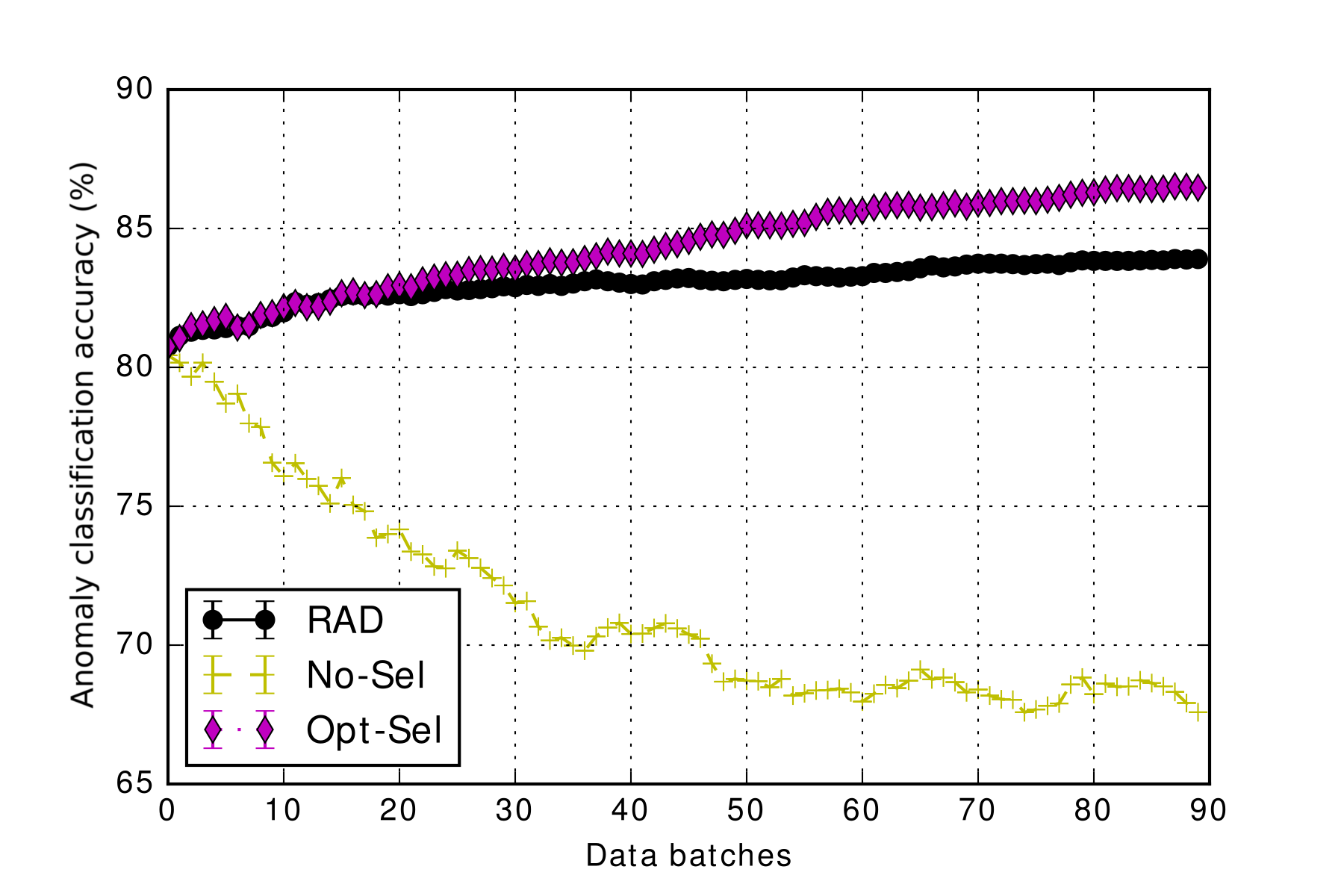}
			\label{fig:EvolOverTime-ClusterTasks-40}
		}
		\caption{Evolution of learning over time -- Use case of Cluster task failures}
		\label{fig:EvolOverTime-ClusterTasks}
	\end{center}
\end{figure*}

\system is developed in Python using scikit-learn~\cite{scikit-learn}. The main performance evaluation metric is accuracy. Experiments are carried out 3 times, results are aggregated by computing mean.

{\bf Noise.} We inject noise into the two datasets by exchanging the true label of data instances with a random one. The label noise is symmetric, i.e., following the noise completely at random model~\cite{Frenay:TNNLS14:survey} where a label is picked with equal probability from all classes except the true one. The noise level $\tilde{Y}$ represents the percentage of data instances with noisy labels. We emulate time-varying noise by drawing for each new data batch the noise level from a Gaussian distribution with 20\% standard deviation and the targeted mean level. We assume that all data is affected by label noise, except the testing data.


{\bf Continual learning.} We start with an initial data batch of 6000 data instances for the Cluster task failures and the IoT devices dataset.
Then, data instances arrive continuously in batches of 600 (Cluster) and 300 (IoT) data instances. Both the initial and subsequent data batches are affected by noise. To kick-start the label and classification models in \system we assume to know which initial data instances are affected by noise (no assumptions for the subsequent data batches). We select 6000 clean data instances as the test dataset for both use case. Test dataset will be used at the end of each epoch to evaluate the accuracy of the trained classification model. We show the evolution of the model accuracy over data batch arrivals until the performance of \system converges.


{\bf Label model.} We use a multilayer perceptron to assess the quality of each label. For IoT and Cluster dataset, the neural network consists of two layers with 28 neurons each. The precision and robustness of the label model are critical to filter out the noisy/malicious labels and provide a clean training set to the classification model. We considered different models, but neural networks provided the best results in terms of accuracy and stability over time. Adaboost gave excellent accuracy when training from the initial data with ground truth, but resulted too sensitive to the unknown noise of subsequent data batches. Random forest is also known to be robust against label noise~\cite{Frenay:TNNLS14:survey}, however its accuracy was below the neural network one.


{\bf Classification model.} We use KNN to assign the correct class label to each data instance filtered by the label model. We set the number of neighbours to five.
Higher values can increase the resilience of the algorithm to residual noise, but also induce extra computational cost. The current choice stems from good results in preliminary experiments.


{\bf Slimmed framework.} For the face recognition task we use \systemSlimmed. In this case we use ResNet~\cite{He2015DeepRL} as classification model. ResNet is a type of CNN architectures which introduces residual functions to alleviate the vanishing gradient problem in training deep neural networks improving the classification performance. 

{\bf Baselines.} The proposed \system is compared against two baseline data selection schemes: (i) {\em No-Sel}, where all data instances of arriving batches are used for training the classification model; and, (ii) {\em Opt-Sel} which emulates an omniscient agent who can perfectly distinguish between clean and noisy labels. The two baselines are representative of the worst and best possible data selection strategies and we expect \system to fall in between. In addition, we consider the {\em Full-Clean} baseline which simulates perfectly recovered labels, i.e., all wrong labels have been correctly identified and recovered by, e.g., an oracle. This represents the ideal solution which provides all clean data in each data batch.

\subsection{Handling Dynamic Data}
\label{ssec:EvalDynamicData}

\begin{figure*}[tb]
	\begin{center}
		\subfloat[IoT thermostat device attacks]{
			\includegraphics[width=0.83\columnwidth]{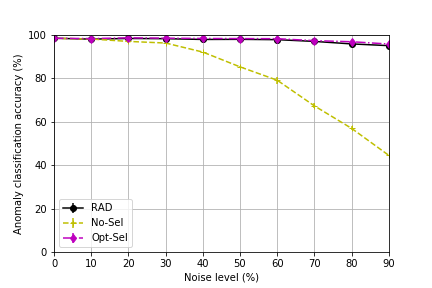}
			\label{fig:eval-noise-IoT}
		}
		\hfil
		\subfloat[Cluster task failures]{
			\includegraphics[width=0.83\columnwidth]{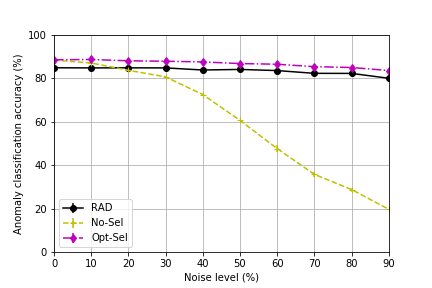}
			\label{fig:eval-noise-ClusterTasks}
		}
		\caption{Impact of data noises on \system accuracy}
		\label{fig:NoiseLevel}
	\end{center}
\end{figure*}

We start by illustrating how \system enables to increase the anomaly detection accuracy over time, despite the presence of noise.
Fig.~\ref{fig:EvolOverTime-IoT} and \ref{fig:EvolOverTime-ClusterTasks} show the evolution of the mean and variance of the classification accuracy achieved by \system  on the thermostat and task failure datasets, respectively.
Each figure moreover presents results under two levels of label noise: 30\% and 40\%. We compare \system against no selection ({\em No-Sel}) and optimal selection ({\em Opt-Sel}).
One can notice that learning from all data instances without cleansing (i.e., {\em No-Sel} curves) gives consistently lower accuracy in all cases. For the attacks classification on the thermostat, the accuracy even oscillates and diverges.
The performance when using \system is better. First because the accuracy does not diverge. Second because it always consistently increase until it saturates. The end accuracies are around 98\% and 84\% for the IoT attack and cluster tasks datasets, respectively.
While for the first dataset the accuracy of \system follows closely the accuracy of {\em Opt-Sel}, for the second dataset \system follows {\em Opt-Sel} at first but then saturates after 40 data batch arrivals.
\system is efficient for various classification applications, however not optimal for all of them.
Note that \system gives also more stable results as shown by shorter errorbars which in magnitude are in line with the ones obtained by an ideal data cleansing. For {\em No-Sel} the bars are significantly larger.
We discuss the results for other IoT devices in Section~\ref{ssec:SystemGenericity}.
\begin{figure}[tb]
	\centering
	\includegraphics[width=0.8\columnwidth]{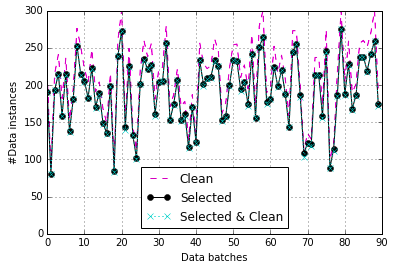}
	\caption{Data selection -- Use case of IoT thermostat device attacks with 30\% noise level.}
	\label{fig:eval-dataselection-IoT}
\end{figure}

Fig.~\ref{fig:eval-dataselection-IoT} presents the variations of noise over time for one run on IoT thermostat dataset where the mean noise rate is set to 30\%. Overlaid is the number of data selected by \system and the overlap between the selected and actually clean data. Results highlight the sharpness of data selection and its parsimony.

In summary: (i) continual learning is advantageous compared to using only the initial dataset; however, (ii) continual learning exposes us to possible classification accuracy degradation stemming from noisy labels if proper data selection is lacking, (iii) \system improves the classification accuracy compared to taking all labels, (iv) the data selection of \system is good, and close to being optimal in some cases.

\subsection{Evaluation of Noise Robustness of \system}
\label{ssec:EvalNoiseImpact}

Next we investigate the impact of different noise levels on the \system performance in terms of classification accuracy.

Fig.~\ref{fig:eval-noise-IoT} and~\ref{fig:eval-noise-ClusterTasks} present the classification accuracy for various levels of noise, ranging from 0\% (all data are clean) up to 90\% for our two main reference datasets: IoT thermostat device attacks and Cluster task failures. Accuracy is measured once the learning has converged. Once again, the \system performance is compared to learning from all data ({\em No-Sel}) and an omniscient data cleanser ({\em Opt-Sel}).

As illustrated in Section~\ref{sec:ProblemStatement}, for {\em No-Sel} the noisier the data are, the worse the classification accuracy, dropping to 20\% and 42\% for the tasks and thermostat datasets, respectively.
A decreasing trend can also be found for \system and {\em Opt-Sel}, however the drops are significantly smaller: at most 5\%. As there is by definition no noise in {\em Opt-Sel} case, the decrease in classification accuracy is only due to the reduction of the overall amount of clean data to learn from. Since the data cleansing of \system is not perfect, the accuracy reduction is caused by noise pollution and overall clean data reduction. Nevertheless, the impact is small and any huge accuracy pitfall is avoided which results in \system's performance being close to {\em Opt-Sel}. We can conclude that \system can limit the impact of the amount of noise across a wide range of noise levels.

\subsection{Analysis of All Datasets}
\label{ssec:SystemGenericity}

Summary results are reported in Table~\ref{tab:table7} and Table~\ref{tab:table8}. In addition to the average accuracy after the last batch arrival, we also underline the accuracy improvement room obtained by comparing {\em Full-Clean} to the {\em No-Sel}, shown in the {\em improvement room} column, and relative accuracy improvement of \system to the {\em No-Sel} strategy, shown in the {\em improvement} column.


\begin{table*}[t]
	\begin{center}
		\caption{Evaluation of the all algorithms for Cluster task failures datasets with 30\% noise level  }
		\label{tab:table7}
		\begin{tabular}{L{3.5cm} C{1.2cm} C{1.2cm}  C{1.2cm} C{1.8cm} C{2cm} C{2cm} C{2cm}}
			\toprule
			Algorithm  & Initial accuracy & No-Sel  & Opt-Sel & Full-Clean & Proposed algorithm & Improvement room & Improvement \\
			\midrule
			\system    &   80.83\%  &  80.92\%  &    87.57\% &  88.43\% &  84.72\% &  7.51\% &  3.8\% \\
			\systemVoting    &   81.12\%  &  80.85\%      &  87.65\% &  88.31\% &  84.0\% &  7.46\% &  3.15\% \\
			\systemActive    &   80.43\%  &  79.56\%    &  87.43\% &  88.27\% &  88.1\% &  8.71\% &  8.54\% \\
			\systemActiveLimit   &   80.87\%  & 79.71\%     &  87.8\% &  87.95\% &  87.35\% &  8.24\% &  7.64\% \\
		\end{tabular}
	\end{center}
\end{table*}

\begin{table*}[t]
	\begin{center}
		\caption{Evaluation of the all algorithms for IoT device attacks datasets with 30\% noise level  }
		\label{tab:table8}
		\begin{tabular}{L{3.5cm} C{1.2cm} C{1.2cm}  C{1.2cm} C{1.8cm} C{2cm} C{2cm} C{2cm}}
			\toprule
			Algorithm  & Initial accuracy & No-Sel  & Opt-Sel & Full-Clean & Proposed algorithm & Improvement room & Improvement \\
			\midrule
			\system    &   94.75\%  &  96.1\%  &    98.35\% &  98.27\% &  98.1\% &  2.17\% &  2.0\% \\
			\systemVoting    &   95.25\%  &  95.32\%      &  98.08\% &  98.47\% &  98.07\% &  3.15\% &  2.75\% \\
			\systemActive    &   95.53\%  &  95.21\%     &  98.23\% &  98.43\% &  98.41\% &  3.22\% &  3.2\% \\
			\systemActiveLimit   &   95.23\%  &  95.67\%    &  98.27\% &  98.5\% &  98.45\% &  2.83\% &  2.78\% \\
		\end{tabular}
	\end{center}
\end{table*}

All results are positive, with varying magnitude depending on the dataset. In all cases, the proposed \system improves between 2\% to 4\% the accuracy compared to blindly taking all data instances. However, more important than the absolute gain is the trend. For example, for the thermostat dataset, we can observe that \system converges over time to a stable level as well as {\em Opt-Sel} model, but {\em No-Sel} diverges. This means that as time goes by, {\em No-Sel} becomes worse and worse.

Even more than the benefits of continual learning might be important the resilience to high levels of noise. Under such levels, the classification accuracy without data cleansing diverges for all datasets. Even if it is rare to have noise levels of 90\% or above, they might still happen for short periods of time in case of attacks to the auto-labelling system via flooding of malicious labels. Hence this property can be crucial for the dependability of the auto-labelling system.

\subsection{Limitation of RAD Framework}
\label{ssec:Limitation}
Though \system works well for datasets of Cluster task failures and IoT device attacks. We can still see the potential limitations of this framework, for example: (1) the assumption of availability of a small fraction of clean data which may not be possible; (2) if data is coming at high rates, or the structure of quality model becomes more complicated, training and prediction time of first layer will slow down the system; (3) as anomaly classifier receives only the data selected by label model, there is a risk that the classifier model overfits to label model. To address these issues we devised two extensions presented in Section~\ref{ssec:extensions} that are evaluated in the next subsections.

\subsection{\systemVoting and History Extension}
\label{ssec:SystemVoting}
\begin{figure*}[t]
	\centering
	\subfloat[Iot data with noise level of 30\%]{
		\includegraphics[width=0.8\columnwidth]{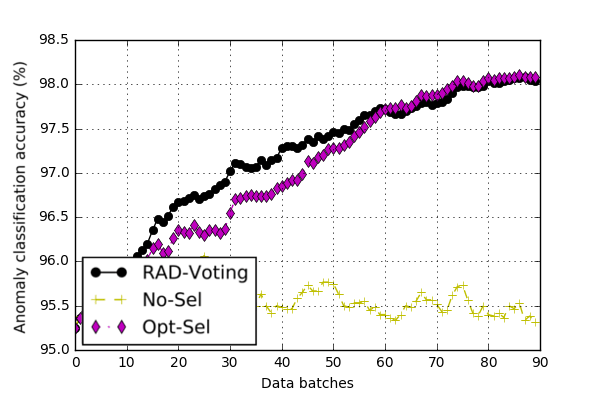}
		\label{fig:EvolOverTime-IoT-Voting-30}
	}
	\hfil
	\subfloat[Iot data with noise level of 40\%]{
		\includegraphics[width=0.8\columnwidth]{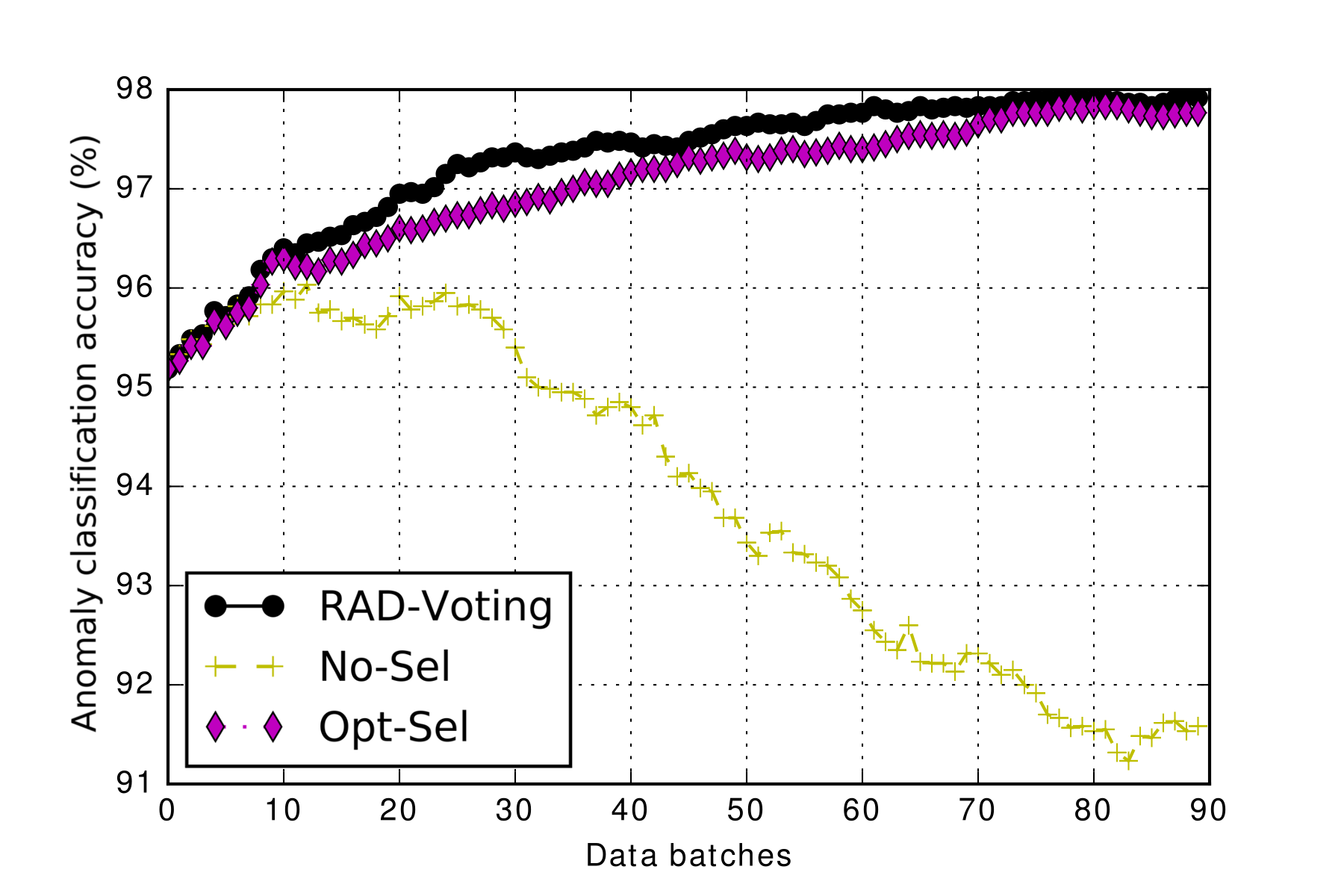}
		\label{fig:EvolOverTime-IoT-Voting-40}
	}
	\caption{Evolution of learning over time -- Use case of IoT thermostat device attacks with \systemVoting}
	\label{fig:EvolOverTime-IoT-voting}
\end{figure*}
\begin{figure*}[!t]
	\centering
	\subfloat[Cluster data with noise level of 30\%]{
		\includegraphics[width=0.8\columnwidth]{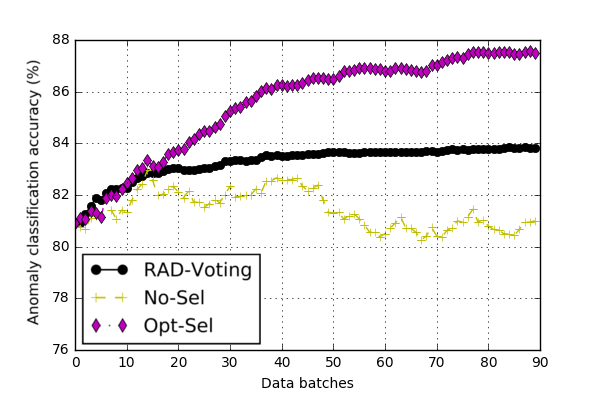}
		\label{fig:EvolOverTime-Cluster-Voting-30}
	}
	\hfil
	\subfloat[Cluster data with noise level of 40\%]{
		\includegraphics[width=0.8\columnwidth]{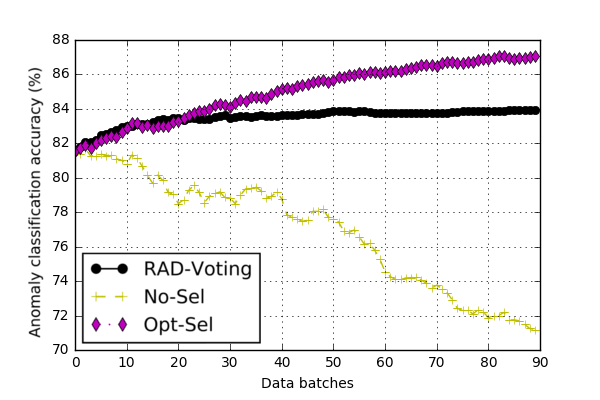}
		\label{fig:EvolOverTime-Cluster-Voting-40}
	}
	\caption{Evolution of learning over time -- Use case of Cluster task failures with \systemVoting}
	\label{fig:EvolOverTime-Cluster-voting}
\end{figure*}
\begin{figure*}[!ht]
	\centering
	\subfloat[Iot data with noise level of 30\%]{
		\includegraphics[width=0.8\columnwidth]{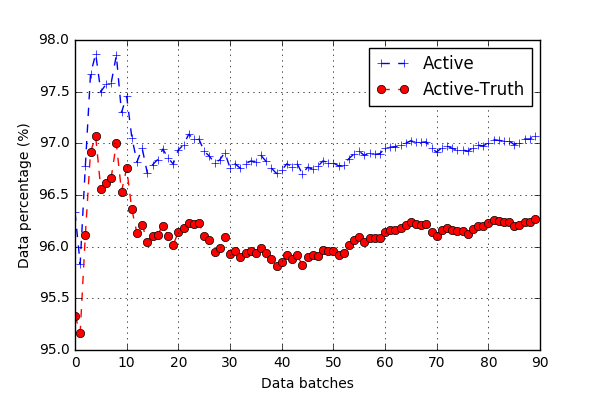}
		\label{fig:EvolOverTime-True-Active-IoT}
	}
	\hfil
	\subfloat[Cluster data with noise level of 30\%]{
		\includegraphics[width=0.8\columnwidth]{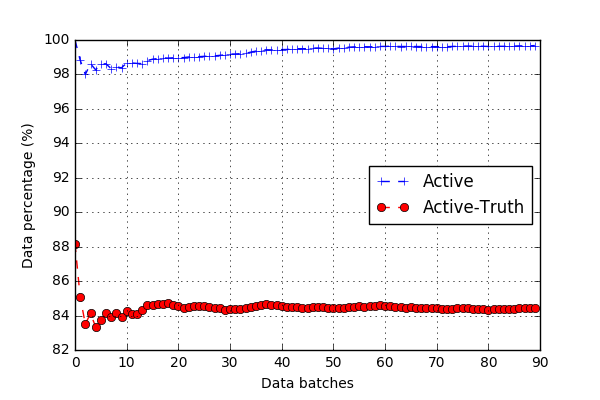}
		\label{fig:EvolOverTime-True-Active-Cluster}
	}
	\caption{Active and active-truth -- voting}
	\label{fig:active-true-active}
\end{figure*}

In the first extension we let both the label and classifier models vote on the label quality and include the possibility to recover instances from history to be evaluated as the models performance improves over time.

We evaluate the accuracy of \systemVoting over time and different noise levels in Fig.~\ref{fig:EvolOverTime-IoT-voting} and Fig.~\ref{fig:EvolOverTime-Cluster-voting} for the IoT thermostat and Cluster task failures, respectively.
We can see that in the case of task failures the \systemVoting performance is similar to the \system performance. However, for IoT dataset, \systemVoting converges faster than \system, and even outperforms {\em Opt-Sel}. This is because we correct labels in \systemVoting algorithm, which increases the number training instances compared to {\em Opt-Sel}. The higher the noise is, the more \systemVoting outperforms {\em Opt-Sel}.

To better understand the different performance between the two data sets let us define
active $A$ as the percent of data used for training till time $t_i$:
\begin{align*}
A = \sum_{k=1}^{i}\frac{ |\mathcal{D}_{k}^{*}| + |\mathcal{U}_k^{*}|}{|\mathcal{D}_k|}.
\end{align*}
Knowing the number of true clean labels used per batch $C_i^T$ we further define active-truth $A^T$ as the percent of true clean active data.
\begin{align*}
A^{T} = \sum_{k=1}^{i}\frac{C_k^T}{|\mathcal{D}_k|}
\end{align*}
In both formulas, we exclude the initial clean batch $\mathcal{D}_0$. Intuitively, $A$ tells how much of the incoming data we use for training, and $A^{T}$ shows how clean the used training data is.

Fig.~\ref{fig:active-true-active}(a) and (b) plot over time these two metrics for the IoT and task failures datasets, respectively.
As seen in Fig.~\ref{fig:active-true-active}(a), for the IoT dataset both $A$ and $A^T$ improve over time. This means that the active data percentage both the amount of active data, i.e. $A$, and the quality of active data $A^T$ improve over time. On the contrary, looking at Fig.~\ref{fig:active-true-active}(b), for the Cluster dataset $A^T$ does not improve over time even if $A$ increases. We impute this to the fact that both $\mathfrak{C}$ and $\mathcal{L}$ predict the same wrong class and this class is used to replace the original label of the data instance.

Table~\ref{tab:table7} and Table~\ref{tab:table8} summarize and compare the  \systemVoting performance to the performance of \system and the other extensions for the Cluster and IoT datasets. 
One thing to note in these two tables is that the values within the columns:  {\em Initial accuracy}, {\em No-Sel}, {\em Opt-Sel} and {\em Full-Clean} should theoretically be the same, respectively. However due to the random initialization of the parameters of machine learning models, there are always minor differences between runs.

\subsection{\systemActive}

\systemActive extends the \system with the ability of asking an oracle to provide the true label for the data instances where the two models do not agree. First we consider \systemActive with no limits on the number of oracle requests followed by \systemActiveLimit which limits the number of oracle interactions.



\begin{figure*}[ht]
	\centering
	\subfloat[Cluster data with noise level of 30\%]{
		\includegraphics[width=0.8\columnwidth]{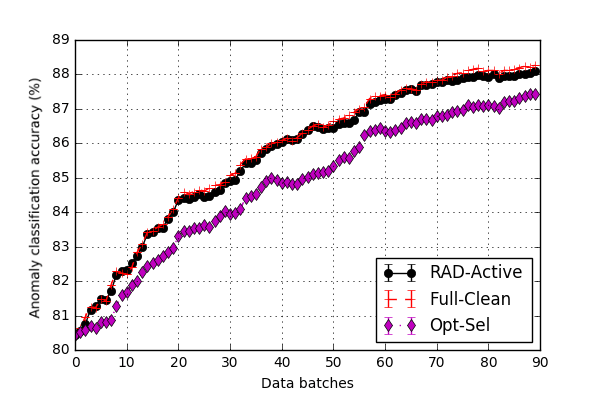}
		\label{fig:EvolOverTime-Cluster-oracle-30}
	}
	\hfil
	\subfloat[Cluster data with noise level of 40\%]{
		\includegraphics[width=0.8\columnwidth]{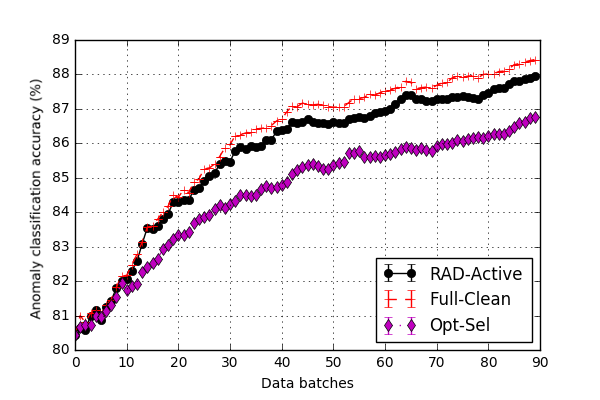}
		\label{fig:EvolOverTime-Cluster-oracle-40}
	}
	\caption{Evolution of learning over time -- active learning}
	\label{fig:EvolOverTime-Cluster-oracle}
\end{figure*}

\begin{figure*}[ht]
	\centering
	\subfloat[Iot data with noise level of 30\%]{
		\includegraphics[width=0.8\columnwidth]{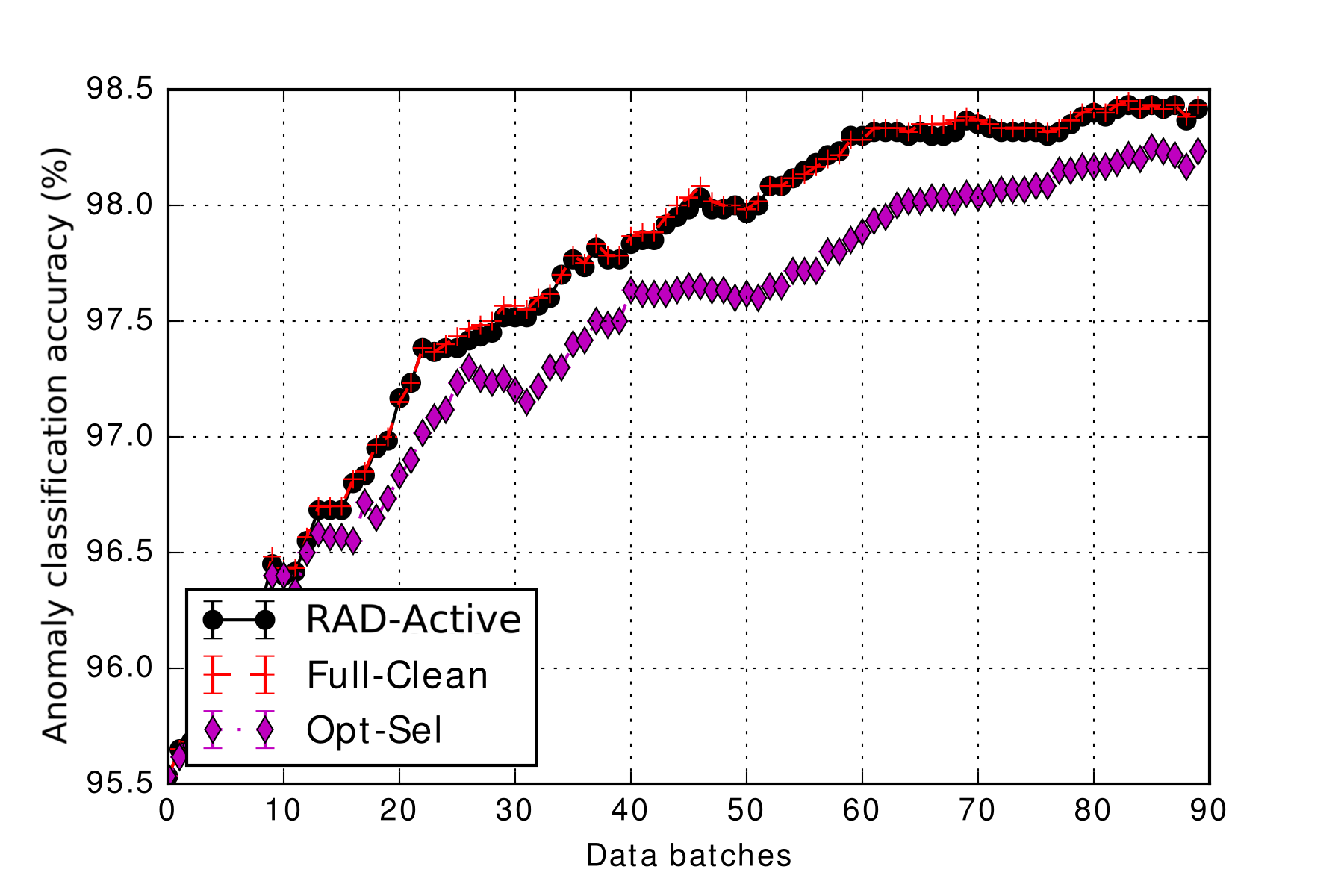}
		\label{fig:EvolOverTime-IoT-oracle-30}
	}
	\hfil
	\subfloat[Iot data with noise level of 40\%]{
		\includegraphics[width=0.8\columnwidth]{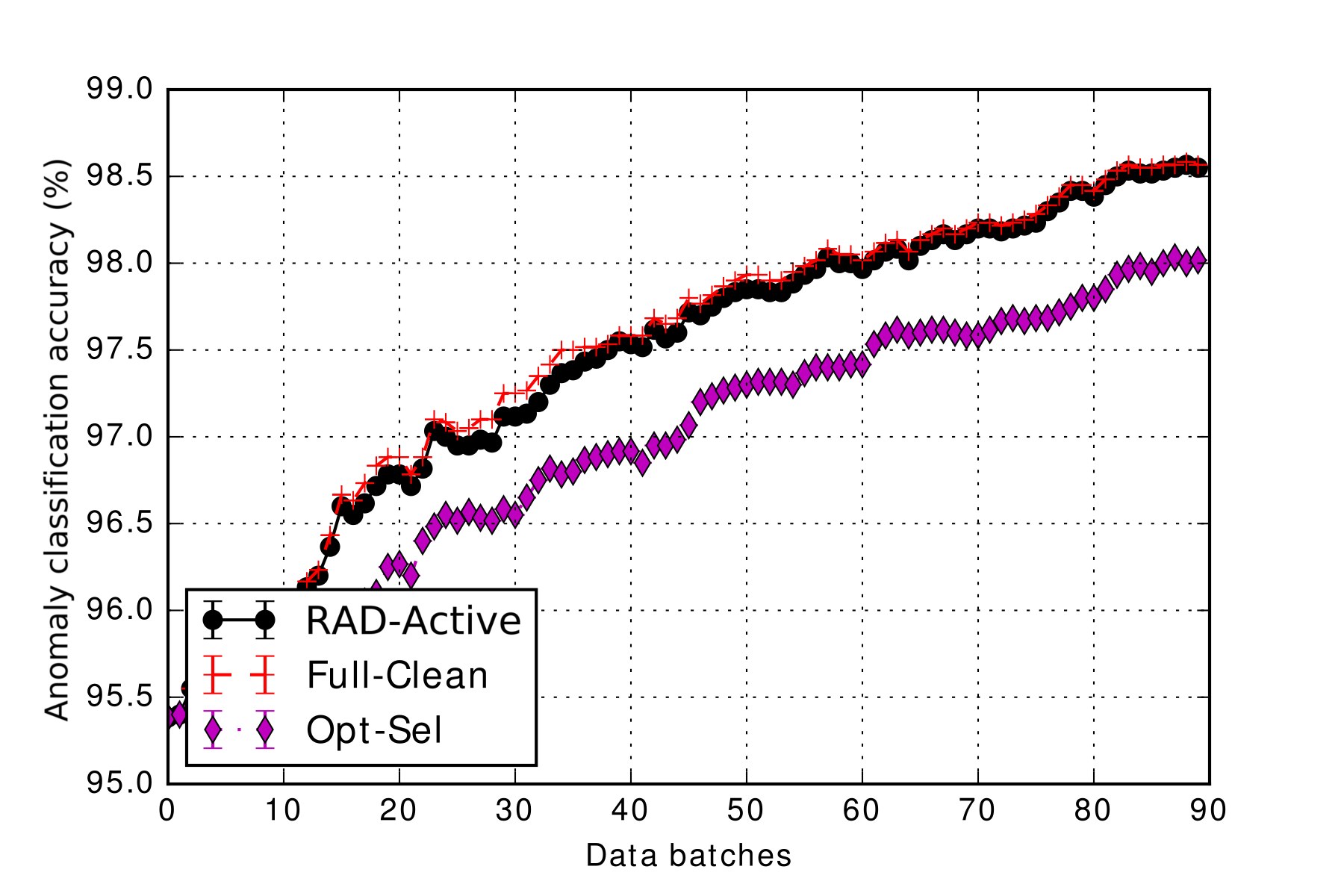}
		\label{fig:EvolOverTime-IoT-oracle-40}
	}
	\caption{Evolution of learning over time -- active learning}
	\label{fig:EvolOverTime-IoT-oracle}
\end{figure*}

Fig.~\ref{fig:EvolOverTime-IoT-oracle} and Fig.~\ref{fig:EvolOverTime-Cluster-oracle} show the performance of \systemActive for the IoT and Cluster datasets, respectively. The figures compare \systemActive to two baselines: {\em Opt-Sel} and {\em Full-Clean}.  We can see that \systemActive is almost as good as {\em Full-Clean} across the two different datasets and different noise levels. Moreover, this is accomplished by consulting the oracle for only about 35\% (30\% noise level) and 39\% (40\% noise level) of the data instances (not counting the initial clean batch $\mathcal{D}_0$).


In reality consulting every single uncertain data instance with expert might be too expensive. Hence we consider \systemActiveLimit which limits the capacity of consultation with expert.
Table~\ref{tab:table7} and Table~\ref{tab:table8} summarize and compare the results of \systemActive and \systemActiveLimit with a 20\% limit for the two datasets. The results show that limiting the consultations degrades slightly the performance compared to \systemActive, but it is still very close to {\em Full-Clean}. This stems from the fact that most data instances we select to send to the oracle are indeed instances affected by label noise and we use the oracle to correct these. Without our framework to filter out the uncertain data, we would not know which data labels are noisy. So we can either give the whole dataset to the oracle to clean up the dataset which likely is expensive, or randomly select a limited number of data instances and acknowledge that some consultations will be wasted on already clean data instances.

\subsection{Impact of Initialization}
\label{ssec:ImpactInitialization}

Here we study the impact on the \system and its extensions of the amount of available ground truth data, i.e.,~the size of the initial dataset $\mathcal{D}_0$.
We vary the number of initial clean data instances from 100 to 6000, and measure the classification accuracy after 90 data batch arrivals. Her we consider the {\em Opt-Sel} baseline since the No-Sel baseline is meant for the framework configuration, not its performance evaluation.

\begin{figure*}[htb]
	\begin{center}
		\subfloat[IoT thermostat device attacks]{
			\includegraphics[width=0.8\columnwidth]{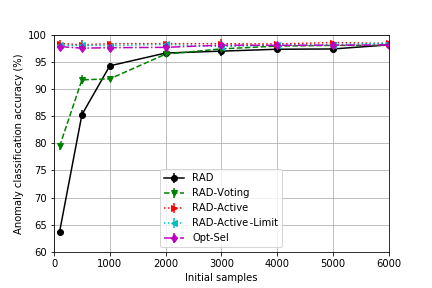}
			\label{fig:eval-init-data-IoT}
		}
		\hfil
		\subfloat[Cluster task failures]{
			\includegraphics[width=0.83\columnwidth]{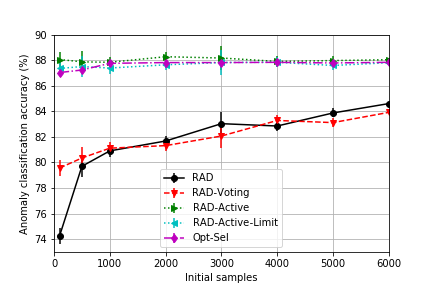}
			\label{fig:eval-init-data-ClusterTasks}
		}
		\caption{Impact of size of initial data batch $\mathcal{D}_0$ on \system accuracy with 30\% noise level}
		\label{fig:IntialCluster}
	\end{center}
\end{figure*}

In Fig.~\ref{fig:eval-init-data-IoT} and Fig.~\ref{fig:eval-init-data-ClusterTasks} show the results for the IoT and Cluster datasets, respectively.

In both figures {\em Opt-Sel} is quite independent from the number of initial data instances ($|D_0|$). This is due to the fact that after 90 batch arrivals the amount of training data is enough to be in the accuracy saturation region. However for \system and \systemVoting, the size of $\mathcal{D}_0$ does matter: the larger the better. At $|\mathcal{D}_0|=3000$ their performances are close to {\em Opt-Sel} for IoT dataset, and at $|\mathcal{D}_0|=6000$ they completely overlap. \systemVoting outperforms \system  in both datasets when $\mathcal{D}_0$ is 100. This is because \systemVoting can correct data labels increasing the number of training instances. Finally, \systemActive and \systemActiveLimit (20\% limit) do not depend on the size of $\mathcal{D}_0$, since they can ask the oracle for the label of uncertain data instances.

This justifies our earlier choice of $D_0$ having 6000 data instances as it enables to achieve the best accuracy. However, all proposed frameworks could also perform well with only half of the data instances in $D_0$.

\subsection{\systemSlimmed on Image Data}
\label{ssec:systemSlimmed}

\begin{figure}[htb]
    \centering
    \includegraphics[width=0.8\linewidth]{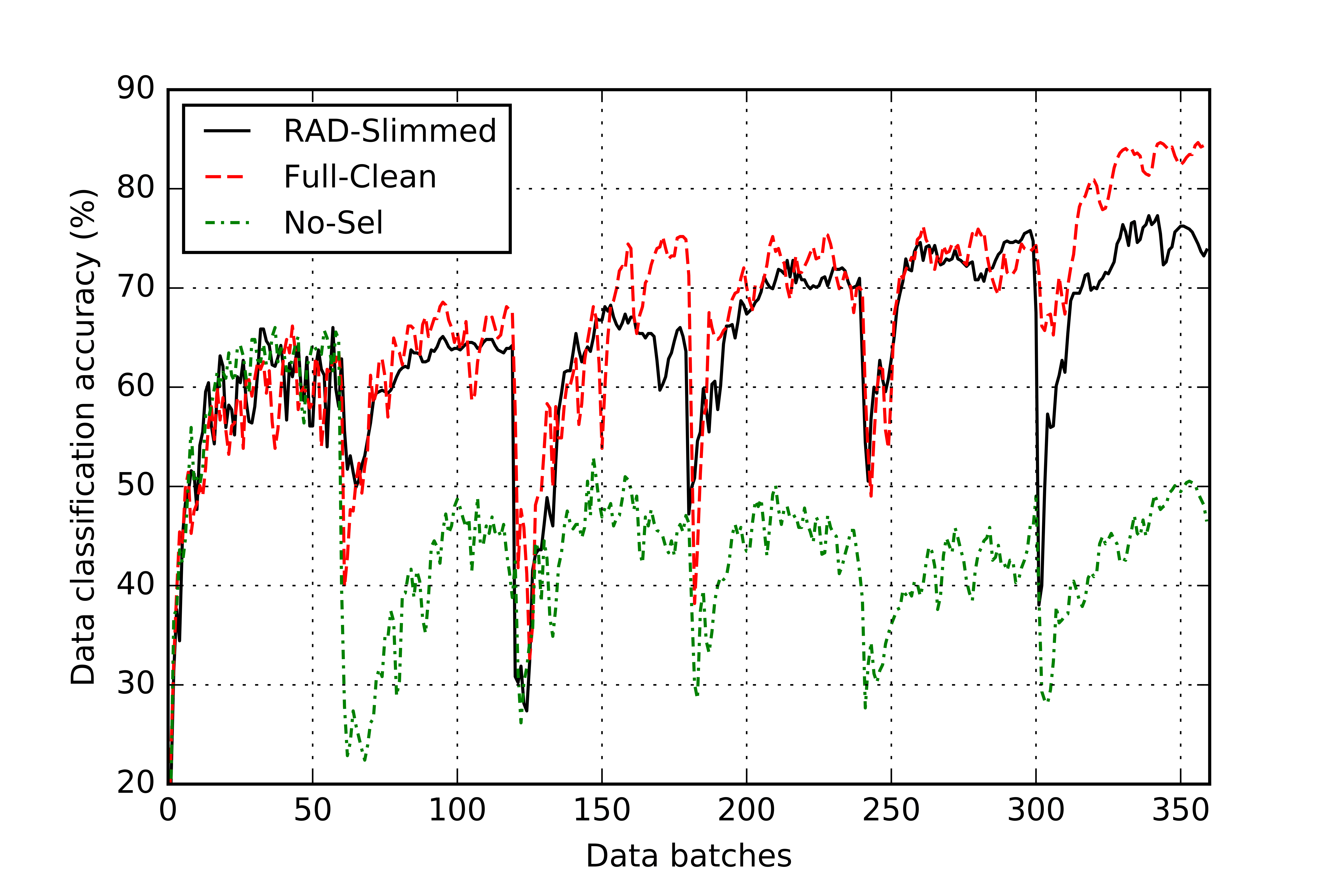}
    \vspace{0.5em}
    \caption{FaceScrub with noise level of 30\%} 
    \label{fig:curve-oracle-image}
\end{figure}

We conclude our evaluation section by testing our framework with image data using  \systemSlimmed to labelling a subset of FaceScrub face images with the celebrity name.
We allow 120 consultations with the oracle for each batch of 400 data instances, i.e., 30\%. Fig.~\ref{fig:curve-oracle-image} shows the accuracy results across the batch arrivals. We can observe that \systemSlimmed is close to the {\em Full-Clean} baseline. Another thing we could notice is that all three curves suffer a periodic up-down pattern. This is because each time a new batch comes, we only use this new batch data as training dataset. As different batches provide different subviews of the data the empirical distribution can be different as well as the calculated optimum, but the model remains. So for the first epoch of a new batch, we will generate a gradient which is based on new data but applied on an old model. This can influence the accuracy of the model. Moreover when retraining on each new data batch we reset the learning rate which causes a bump in the learning rate. Therefore, even if all batches follow the same distribution, the system could temporarily wander off from the previous optimum.

More complete results are showed in Table~\ref{tab:table9} with varying consultation limits. We see that increasing the consultations limit allows the final accuracy to get closer and closer to {\em Full-Clean}.
The difference between {\em No-Sel} and $0\%$ is that {\em No-Sel} uses directly the noisy label data to train the classifier, whereas 0\% uses the classifier to filter the uncertain data instances, but it is not allowed to send it to the oracle for consultation.

\begin{table*}[t]
	\begin{center}
		\caption{Evaluation of the \systemActiveLimit for FaceScrub with different limits per batch}
		\label{tab:table9}
		\begin{tabular}{L{1.2cm} C{1.2cm} C{1.2cm}C{1.2cm}  C{1.2cm} C{1.2cm} C{1.2cm} C{1.2cm} C{1.2cm} C{1.2cm}}
			\toprule
			 Initial accuracy & No-Sel & Opt-Sel & Full-Clean & 0\%& 10\%& 20\%& 30\%\\
			\midrule
			  63.61\%  &  46.47\%  &  80.15\%  & 84.36\% & 48.72\% & 66.32\%&70.68\% &  74.14\%  \\
		\end{tabular}
	\end{center}
\end{table*}

%% file: RelatedWork.tex
\section{Related Work}
\label{sec:RelatedWork}



Machine learning has been extensively used for failure detection~\cite{Pitakrat:2013,Pellegrini:2015,Reuter:2018,Campos:2018},
attack prediction~\cite{Agarwal:2016,Banescu:2017,Anbar:2018,Kozik:2018,Karagiannis:2018,Zhou:2018}, and face recognition~\cite{DBLP:conf/cvpr/TaigmanYRW14,DBLP:conf/cvpr/SchroffKP15,DBLP:conf/cvpr/WangWZJGZL018}.
Considering noisy labels in classification algorithms is also a problem that has been explored in the machine learning community as discussed in~\cite{Frenay:TNNLS14:survey,biggio2011support,natarajan2013learning}.

The problem of classification in presence of noisy labels can be organized into various categories according to, on the one hand, the type of classification algorithm subject to noise, and on the other hand, the techniques used to remove the noise.

Regarding the type of classification algorithm, the problem of noisy labels has been studied both for binary classification where noisy labels are considered as symmetric (e.g., \cite{larsen1998design}) and for classification with multiple classes where noisy labels are considered as asymmetric, e.g.,~\cite{patrini2017making,sukhbaatar2014training}. In the context of this paper, we consider the problem of classification with multiple classes.
Furthermore, noisy labels have been considered in various types of classifiers KNN~\cite{Wilson:ML2000:MLselection}, SVM~\cite{An:Neurocomput.13:SVM}, and deep neural networks~\cite{Vahdat:NIPS17:NoisyLabelDNN}. In the context of this paper, our proposed approach is agnostic to the underlying classifier type as noise removal is performed ahead of the classification.

To deal with noise, various techniques have been explored including forward loss correction.
These algorithms learn about the label noise by adjusting the loss to the end of the model. However, these solutions either rely on strong assumptions or have limited accuracy as they generally do not rely on clean labels to remove the noise. More accurate solutions, which rely on clean labels during the training phase have thus been explored (e.g., \cite{veit2017learning,li2017learning,Dan:nips2018:Using}). These solutions generally train a separate network for distinguishing noisy labels from clean ones. Robustness to label noise has also been studied for GANs performing image recognition, both in the context of known and unknown noise distribution \cite{thekumparampil2018robustness}. However, all these solutions have been designed and tested on static datasets and in an off-line setting. Instead in the context of this paper, we consider a dynamic model where the network has been trained using cleaned labels continues to learn over time.

%% file: Conclusion.tex
\section{Concluding Remarks}
\label{sec:Conclusion}

While machine learning classification algorithms are widely applied to detect anomalies, the commonly employed assumption of clean anomaly labels often does not hold for the data collected in the wild due to careless annotation and malicious dirty label pollution. The noisy labels can significantly degrade the accuracy of anomaly detection with an increasing amount of data  and are challenging to tackle due to the lack of ground truth of label quality. In this paper, we present a on-line framework for robust anomaly detection, \system, which can continuously learn the system dynamics and anomaly behaviours from streams of arriving data after filtering out suspicious noisy data. 

\system is a general framework that composes of sequence of quality and classification models, where the former captures the label dynamics and the latter focus on detection anomaly. To overome the limtations of on-line learning,  the additional features of \system are repeatitively cleaninsing and oracle queries which are based on the learning capacity varying over time.  We demonstrate the effectiveness of \system on three uses cases, i.e., detecting IoT device attacks, predicting task failure at Google clusters and recognising celebrity faces using FaceScrub. \system can robustly improve the detection accuracy against different levels of label noises, reaching up to 74\%, 83\% and 98\% accuracy for recognising face, predicting task failure and detecting IoT device attacks, respectively, whereas  learning directly from all the data streams without filtering degrades the detection accuracy. 

